\documentclass[10pt,twocolumn,letterpaper]{article}

\usepackage{cvpr}              
\usepackage[accsupp]{axessibility}
\usepackage{makecell}
\usepackage{bm}
\usepackage{multirow}
\usepackage{utfsym}
\usepackage{makecell}
\usepackage{bm}
\usepackage{colortbl}
\usepackage{bbding}
\usepackage{booktabs}
\usepackage{utfsym}
\usepackage{graphicx}
\usepackage{amsmath}
\usepackage{amssymb}
\usepackage{xcolor}
\usepackage{pifont}
\usepackage{diagbox}
\usepackage[linesnumbered,ruled,vlined]{algorithm2e}

\definecolor{cvprblue}{rgb}{0.21,0.49,0.74}
\usepackage[pagebackref,breaklinks,colorlinks,citecolor=cvprblue]{hyperref}

\def\paperID{6926} 
\def\confName{CVPR}
\def\confYear{2024}

\title{Enhance Image Classification via Inter-Class Image Mixup with Diffusion Model
}
\author{Zhicai Wang$^1$\thanks{This work was done during the internship in Huawei Inc..}, Longhui Wei$^{2 \dagger}$, Tan Wang$^3$, Heyu Chen$^1$, Yanbin Hao$^1$, Xiang Wang$^{1}$\thanks{Xiang Wang and Longhui Wei are both the corresponding authors.},\\ Xiangnan He$^1$, Qi Tian$^2$ \\
$^1$University of Science and Technology of China, $^2$ Huawei Inc.,\\ $^3$ Nanyang Technological University\\}

\begin{document}
\maketitle

\begin{abstract}
    Text-to-image (T2I) generative models have recently emerged as a powerful tool, enabling the creation of photo-realistic images and giving rise to a multitude of applications. However, the effective integration of T2I models into fundamental image classification tasks remains an open question. A prevalent strategy to bolster image classification performance is through augmenting the training set with synthetic images generated by T2I models. In this study, we scrutinize the shortcomings of both current generative and conventional data augmentation techniques. Our analysis reveals that these methods struggle to produce images that are both faithful (in terms of foreground objects) and diverse (in terms of background contexts) for domain-specific concepts. To tackle this challenge, we introduce an innovative inter-class data augmentation method known as Diff-Mix \footnote{\href{https://github.com/Zhicaiwww/Diff-Mix}{https://github.com/Zhicaiwww/Diff-Mix}}, which enriches the dataset by performing image translations between classes. Our empirical results demonstrate that Diff-Mix achieves a better balance between faithfulness and diversity, leading to a marked improvement in performance across diverse image classification scenarios, including few-shot, conventional, and long-tail classifications for domain-specific datasets.
\end{abstract}



\begin{figure}
    \includegraphics[width=1\linewidth]{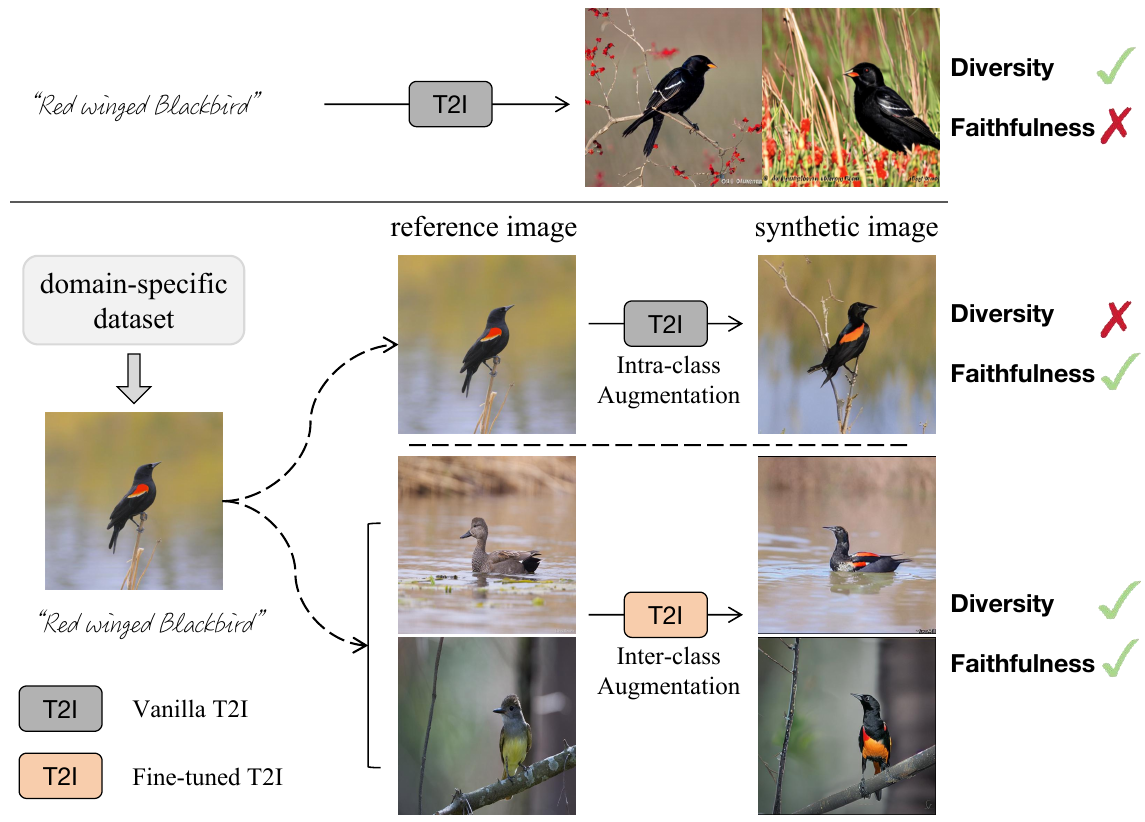}
    \caption{Strategies to expand domain-specific datasets for improved classification are varied. Row 1 illustrates vanilla distillation from a pretrained text-to-image (T2I) model, which carries the risk of generating outputs with reduced faithfulness. Intra-class augmentation, depicted in Row 2, tends to yield samples with limited diversity to maintain high fidelity to the original class. Our proposed method, showcased in Rows 3 and 4, adopts an inter-class augmentation strategy. This involves introducing edits to a reference image using guidance of another class, which significantly enriches the dataset with a greater diversity of samples.
\label{fig:introduction}}
\vspace*{-0.3cm}
\end{figure}

\section{Introduction}
In comparison to GAN-based models \cite{goodfellow2014generative,stylegan,biggan}, contemporary state-of-the-art text-to-image (T2I) diffusion models exhibit enhanced capabilities in producing high-fidelity images \cite{beatsGan,Imagen,glide,LatentDiffusion}.
With the remarkable cross-modality alignment capabilities of T2I models, there is significant potential for generative techniques to enhance image classification \cite{SyntheticImageNet,GenerativeRoubustness}.
%
For instance, a straightforward approach entails augmenting the existing training dataset with synthetic images generated by feeding categorical textual prompts to a T2I diffusion model. However, upon reviewing prior approaches employing T2I diffusion models for image classification, it becomes evident that the challenge in generative data augmentation for domain-specific datasets is producing samples with both a \textbf{faithful foreground} and a \textbf{diverse background}.
Depending on whether a reference image is used in the generative process, we divide these methods into two groups:
\begin{itemize}
    \item Text-guided knowledge distillation \cite{FakeImageNet-SD, tian2023stablerep} involves generating new images from scratch using category-related prompts to expand the dataset. For the off-the-shelf T2I models, such vanilla distillation presume these models have comprehensive knowledge of target domain, which can be problematic for domain-specific datasets.  Insufficient domain knowledge easily makes the distillation process less effective. For example, vanilla T2I models struggle to generate images that accurately represent specific bird species based solely on their names (see Row 1 of Fig. \ref{fig:introduction}).
    
    \item Generative data augmentation \cite{antoniou2017data, zheng2024toward} employs generative models to enhance existing images. Da-fusion \cite{dafusion}, for instance, translates the source image into multiple edited versions within the \textbf{same class}. This strategy, termed \textbf{intra-class} augmentation, primarily introduces intra-class variations. While intra-class augmentation retains much of the original image's layout and visual details, it results in limited background diversity (see Row 2 of Fig. \ref{fig:introduction}). However, synthetic images with constrained diversity may not sufficiently enhance the model's ability to discern foreground concepts.
    
    
\end{itemize}
Based on these observations, a fundamental question emerges: `\textit{Is it feasible to develop a method that optimizes both the diversity and faithfulness of synthesized data simultaneously}?'

In this work, we introduce \textbf{Diff-Mix}, a simple yet effective data augmentation method that harnesses diffusion models to perform \textbf{inter-class} image interpolation, tailored for enhancing domain-specific datasets.
The method encompasses two pivotal operations: personalized fine-tuning and inter-class image translation. Personalized fine-tuning \cite{ruiz2023dreambooth, textualinversion} is originally designed for customizing T2I models and enabling them to generate user-specific contents or styles. In our case, we implement the technique to tailor the model, enabling it to generate images with faithful foreground concepts.
Inter-class image translation in Diff-Mix entails transforming a reference image into an edited version that incorporates prompts from different classes. This translation strategy is designed to retain the original background context while editing the foreground to align with the target concept. For instance, as depicted in the bottom rows of Fig. \ref{fig:introduction}, Diff-Mix can generate images of land birds in diverse settings, such as maritime environments, enriching the dataset with a variety of counterfactual samples.

Unlike previous non-generative augmentation methods, such as Mixup \cite{zhang2017mixup} and CutMix \cite{yun2019cutmix}, Diff-Mix works in a foreground-perceivable inter-class interpolation manner and shares a different mechanism with the non-generative approaches. 
Our experiment under the conventional classification setting indicates that incorporating both CutMix and Diff-Mix could further enhance performance. Additionally, when compared with other generative approaches, we conduct experiments under few-shot and long-tail scenarios and observe consistent performance improvements. Our contributions can be summarized as follows:

\begin{itemize}
    \item We pinpoint the critical factors that affect the efficacy of generative data augmentation in domain-specific image classification: namely, faithfulness and diversity.
    \item We introduce Diff-Mix, a simple yet effective generative data augmentation strategy that leverages fine-tuned diffusion models for inter-class image interpolation.
    \item We conduct a comparative analysis of Diff-Mix with other distillation-based and intra-class augmentation methods, as well as non-generative approaches, highlighting its unique features and benefits.
\end{itemize}

\section{Related Works}
\noindent{\textbf{Text-to-image diffusion models.}} 
Following pretraining on web-scale data, the T2I diffusion model has demonstrated robust capabilities in generating text-controlled images \cite{glide, Imagen, LAION, sun2023inner}.
Its versatility has led to diverse applications, including novel view synthesis \cite{watson2022novel, bhunia2023person}, concept learning \cite{customdiffusion, ruiz2023dreambooth}, and text-to-video generation \cite{makeavideo, ImagenVideo}, among others. Recent advancements \cite{zero-shotClassifier} also highlight the cross-modality features of such generative models, showcasing their ability to serve as zero-shot classifiers. 

\vspace{5pt}
\noindent{\textbf{Synthetic data for image classification.}}
There are two perspectives on the utilization of synthetic data for image classification: knowledge distillation \cite{SyntheticImageNet} and data augmentation \cite{beckham2019adversarial, sandfort2019data, shipard2023boosting, wang2023bi}. From the knowledge distillation perspective, SyntheticData \cite{SyntheticData} reports significant performance gains in both zero-shot and few-shot settings by leveraging off-the-shelf T2I models to obtain synthetic data. The work of \cite{SyntheticImageNet} has indicated that fine-tuning the T2I model on ImageNet \cite{imagenet} yields improved classification accuracy by narrowing the domain gap. Some works also find that learning from the synthetic data presents strong transferability \cite{tian2023stablerep, SyntheticData} and robustness \cite{GenerativeRoubustness,IGV,bc_loss}. 
From the data augmentation standpoint, Da-fusion \cite{dafusion} achieves stable performance improvements on few-shot datasets by augmenting from reference images. In a related study \cite{bai2023improving}, the use of StyleGAN \cite{stylegan} for generating interpolated images between two different domains has been shown to enhance classifier robustness for out-of-distribution data. Our work shares similarities with AMR \cite{beckham2019adversarial}, which generates realistic novel examples by interpolating between two images using GAN \cite{gan}. The distinction lies in our discussion of interpolation using the T2I diffusion model, where its noise-adding and denoising characteristics enable a smoother implementation of interpolation.

\vspace{5pt}
\noindent{\textbf{Non-generative data augmentation.}}
Mixup \cite{zhang2017mixup} and CutMix \cite{yun2019cutmix} stand out as two prominent non-generative data augmentation methods, serving as effective regularization techniques during training. While Mixup achieves augmented samples through a convex combination of two images, CutMix achieves augmentation by cutting and pasting parts of images. However, both methods are constrained in their ability to produce realistic images. In addressing this limitation, the utilization of generative models emerges as a potential solution to alleviate this issue.

\section{Method}

\subsection{Preliminary}
\noindent\textbf{Text-to-Image diffusion model.} Diffusion models generate images by gradually removing noise from a Gaussian noise source \cite{ddpm}. In a diffusion process with a total of $T$ steps, its forward process, which gradually adds noise, is represented as a Markov chain with a Gaussian transition kernel, where $q(\bm x_{t}|\bm x_{t-1}) = \mathcal{N}\left(\boldsymbol{x}_t ; \sqrt{\alpha_t} \boldsymbol{x}_{t-1}, (1-\alpha_t) \mathbf{I}\right)$,
where $x_t$ represents the noisy image at step $t$. The training objective at step $t$ is to predict the noise to reconstruct $\bm x_{t-1}$. When training a text-conditioned diffusion model, the simplified training objective can be summarized as follows:
\begin{align}
\mathbb{E}_{\epsilon, \mathbf{x}, \mathbf{c}, t}\left[\left\|\epsilon - \epsilon_\theta\left(\mathbf{x}_t, \mathbf{c}, t\right)\right\|_2^2\right],
\label{eq: training objective}
\end{align}
where $\epsilon_\theta$ represents the predicted noise, and $\bm c$ is the encoded text caption associated with the image $\bm x$.

\vspace{5pt}
\noindent\textbf{T2I personalization.} T2I personalization aims to personalize a diffusion model for generating specific concepts using a limited number of concept-oriented images \cite{customdiffusion, qiu2024controlling, ConceptBed}. These concepts are typically represented using identifiers (\eg, ``\texttt{[V]}"). As a result, we formalize the constructed image-caption set as {$\bm x$, ``\texttt{photo of a [V]}"}. Various personalization methods differ in their fine-tuning strategies. For instance, Textual Inversion (TI) \cite{textualinversion} makes the identifier learnable, but other modules are not fine-tuned, potentially sacrificing some faithfulness in image generation. On the other hand, Dreambooth (DB) \cite{ruiz2023dreambooth} fine-tunes the U-Net \cite{UNet} for more refined personalized generation but faces the challenge of increased computational cost. 

\vspace{5pt}
\noindent\textbf{Image-to-image translation.} Image translation enables image synthesis and editing using a reference image as guidance \cite{isola2017image, cycleGAN, wu2021stylespace}. Diffusion-based image translation methods can generate fine edits, which refer to subtle modifications, with varying degrees of shift relative to the reference image \cite{meng2021sdedit,instructpix2pix,plugandplayTranslation}. Here, we draw inspiration from SDEdit \cite{meng2021sdedit} to perform edits on the reference image, where the target image $\bm x^{\text{tar}}$ is translated from a reference image $\bm x^{\text{ref}}$. During translation, the reverse process does not traverse the full process but starts from a certain step $\lfloor sT \rfloor$, where $s\in[0,1]$ controls the insertion position of the reference image with noise, as follows,
\begin{align}
    \bm x_{\left\lfloor sT \right\rfloor} = \sqrt{\tilde{\alpha}_{\left\lfloor sT \right\rfloor}} \bm x_0^{\mathrm{ref}} + \sqrt{1-\tilde{\alpha}_{\left\lfloor sT \right\rfloor}} \epsilon.
    \label{eq: SDEdit}
\end{align}
By adjusting the strength parameter $s$, one can strike a balance between the diversity of the generated images and their faithfulness to the reference image.

\subsection{General Framework}
The Diff-Mix pipeline consists of two key steps. Firstly, to produce more faithful images for domain-specific datasets, we propose treating it as a T2I personalization problem and fine-tuning the Stable Diffusion (SD). Subsequently, to enhance the diversity of synthetic data beyond the well-fitted training distribution, we employ inter-class image translation. This process produces interpolated images with increased background diversity for each class.

\subsection{Fine-tune Diffusion Model}

Vanilla distillation tends to be less effective, especially as the number of training shots increases (refer to Sec. \ref{sec:low-shot classification}). In order to mitigate the distribution gap, we propose fine-tuning Stable Diffusion in conjunction with current widely-used T2I personalization strategies.
\begin{figure}
    \begin{subfigure}{1\linewidth}
        \includegraphics[width=1\linewidth]{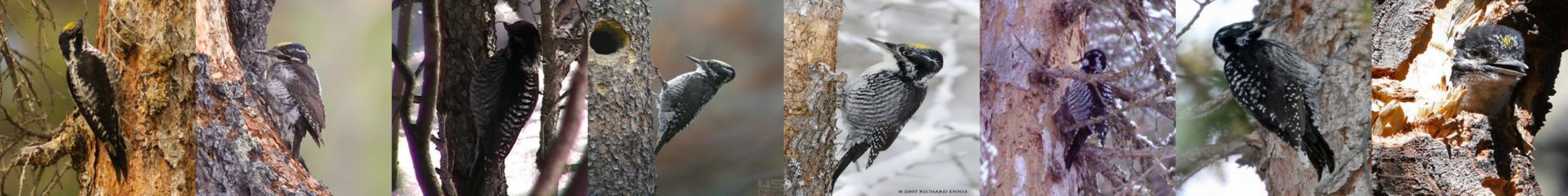}
        \caption{Real images}
    \end{subfigure}

    \begin{subfigure}{1\linewidth}
        \includegraphics[width=1\linewidth]{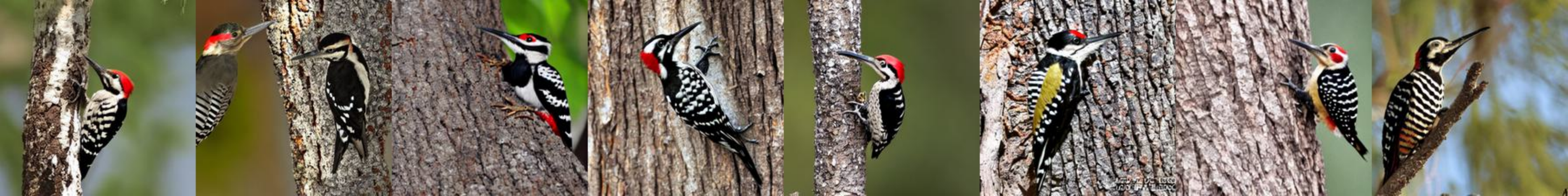}
        \caption{Synthetic images (vanilla SD)}
        \end{subfigure}

    \begin{subfigure}{1\linewidth}
        \includegraphics[width=1\linewidth]{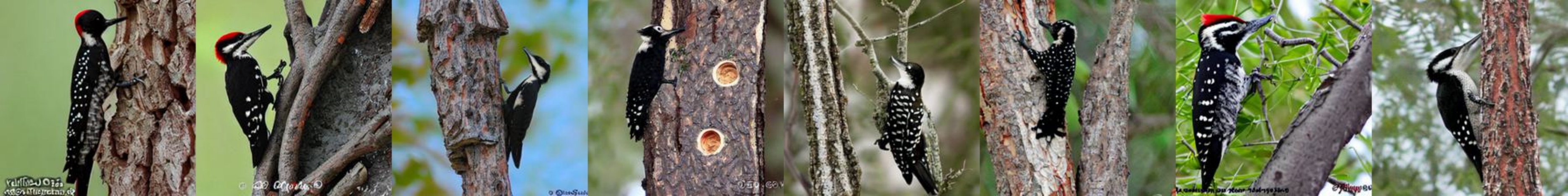}
        \caption{Synthetic images (SD fine-tuned via DB)}
        \end{subfigure}
    \begin{subfigure}{1\linewidth}
        \includegraphics[width=1\linewidth]{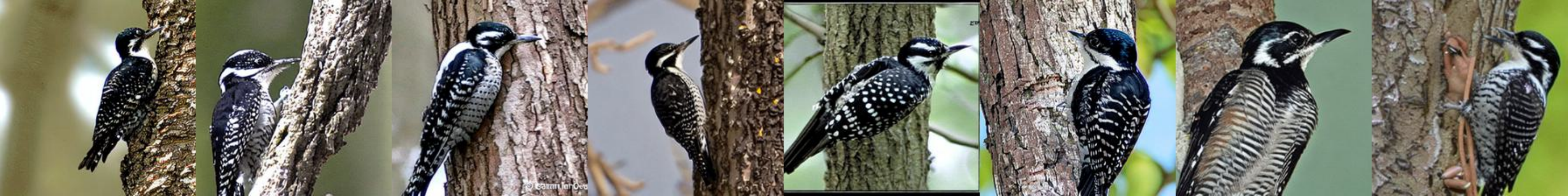}
        \caption{Synthetic images (SD fine-tuned via TI+DB)}
    \end{subfigure}
    \caption{Examples of ``\texttt{American Three toed Woodpecker}''. (a) Real images from the training set.  (b-d) synthetic images generated using different fine-tuned models with the same number of fine-tuning steps. TI+DB indicates both text embedding and U-Net are fine-tuned. TI+DB achieves a more faithful output compared to DB alone (check the head and wing patterns of the birds). \label{fig: case_finetune_strategy}}
    \vspace*{-0.2cm}
\end{figure}

\begin{figure}
  \centering
    \includegraphics[width=1\linewidth]{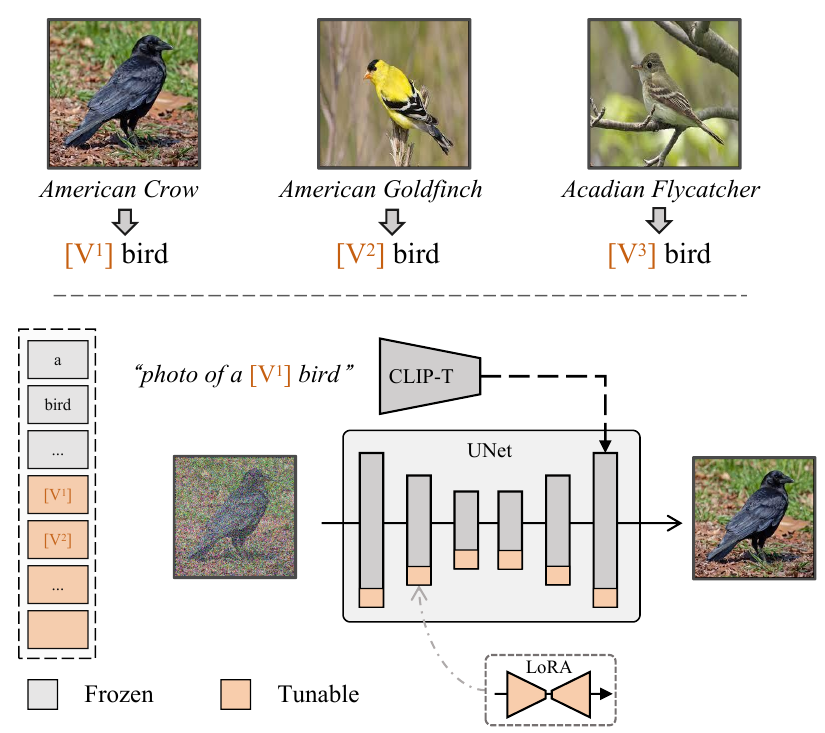}
    \caption{Fine-tuning framework of Diff-Mix operates as follows: Initially, we replace the class name with a structured identifier formatted as ``\texttt{[V$^i$] [metaclass]}'', thereby sidestepping the need for specific terminological expressions. Next, we engage in joint fine-tuning of these identifiers and the low-rank residues (LoRA) of U-Net to capture the domain-specific distribution. \label{fig:method_finetune_strategy}}
    \vspace{-0.3cm}
\end{figure}

\vspace{5pt}
\noindent\textbf{Dreambooth meets Textual Inversion.}
Many fine-grained datasets provide terminological names for their categories, like ``\texttt{American Three toed Woodpecker}'' and ``\texttt{Pileated Woodpecker}''. We could construct image-text pairs using category-related prompts and fine-tune the denoising network of SD using Eq. \ref{eq: training objective}, which is analogous to Dreambooth. However, we observe that directly incorporating these specialized terms into the text during fine-tuning can impede convergence and hinder the generation of faithful images. We attribute this challenge to the semantic proximity of terminology within a fine-grained domain, where fine-tuning the vision module alone tends to be less effective at distinguishing two similar classes within the same family, like ``\texttt{Woodpecker}''.
Inspired by Textual Inversion \cite{textualinversion}, we opt to replace the terminological name in the dataset with ``\texttt{[V$^i$] [metaclass]}'' where ``\texttt{[V$^i$]}" is a learnable identifier, and $i$ varies from $1$ to $N$, representing the category index. The illustration of our fine-tuning strategy is presented in Fig.  \ref{fig:method_finetune_strategy}.
The term ``\texttt{[metaclass]}" is determined by the theme of the current dataset, such as ``\texttt{bird}" for a fine-grained bird dataset. By concurrently fine-tuning the identifier and the U-Net, we empower the model to quickly adapting to the fine-grained domain, allowing it to generate faithful images using the identifier (see comparison between Row 3 and Row 4 in Fig. \ref{fig: case_finetune_strategy}).

\vspace{5pt}
\noindent\textbf{Parameter efficient fine-tuning.}
In this context, we embrace the parameter-efficient fine-tuning strategy known as LoRA \cite{hu2021lora}. LoRA distinguishes itself by fine-tuning the residue of low-rank matrices instead of directly fine-tuning the pre-trained weight matrices. To elaborate, consider a weight matrix $\bm W \in \mathbb{R}^{m\times n}$. The tunable residual matrix $\Delta\bm W$ comprises two low-rank matrices: $\bm A \in \mathbb{R}^{m\times d}$ and $\bm B \in \mathbb{R}^{n\times d}$, defined as $\Delta\bm W = \bm A \bm B^\top$. As a default configuration, we set the rank $d$ to 10. 

\subsection{Data Synthesis Using Diffusion Model}
\label{sec:method_data_synthesis}

In generating pseudo data, three strategies can be used with our fine-tuned diffusion model \footnote{We use the prefix ``Diff-'' denotes the T2I model is fine-tuned and ``Real-'' denotes the vanilla T2I model.}: (1) distillation-based method \textbf{Diff-Gen}, (2) intra-class augmentation \textbf{Diff-Aug}, and (3) inter-class augmentation \textbf{Diff-Mix}. 

        

\vspace{5pt}
\noindent\textbf{Diff-Gen and Diff-Aug.} For a target class $y^i$ and its textual condition, ``\texttt{photo of a \textup{[$\text{V}^i$] [metaclass]}}", both methods generate synthetic samples annotated with class $i$. Specifically, Diff-Gen generates samples from scratch by initializing with random Gaussian noise and proceeding through the full reverse diffusion process with $T$ steps. Diff-Gen can produce images aligned with its fine-tuned distribution. In contrast, Diff-Aug sacrifices a portion of diversity and generate images by editing on a reference image. Specifically, it randomly sample a image from the intra-class training set and enhances the image through image translation using Eq. \ref{eq: SDEdit}. The term ``intra-class" means that the conditioning prompts are constructed based on the ground truth categories of images, and such a denoising process tends to introduce less variation, particularly for the foreground concepts (see top rows of Fig. \ref{fig:introduction}).

\begin{figure}
  \includegraphics[width=1\linewidth]{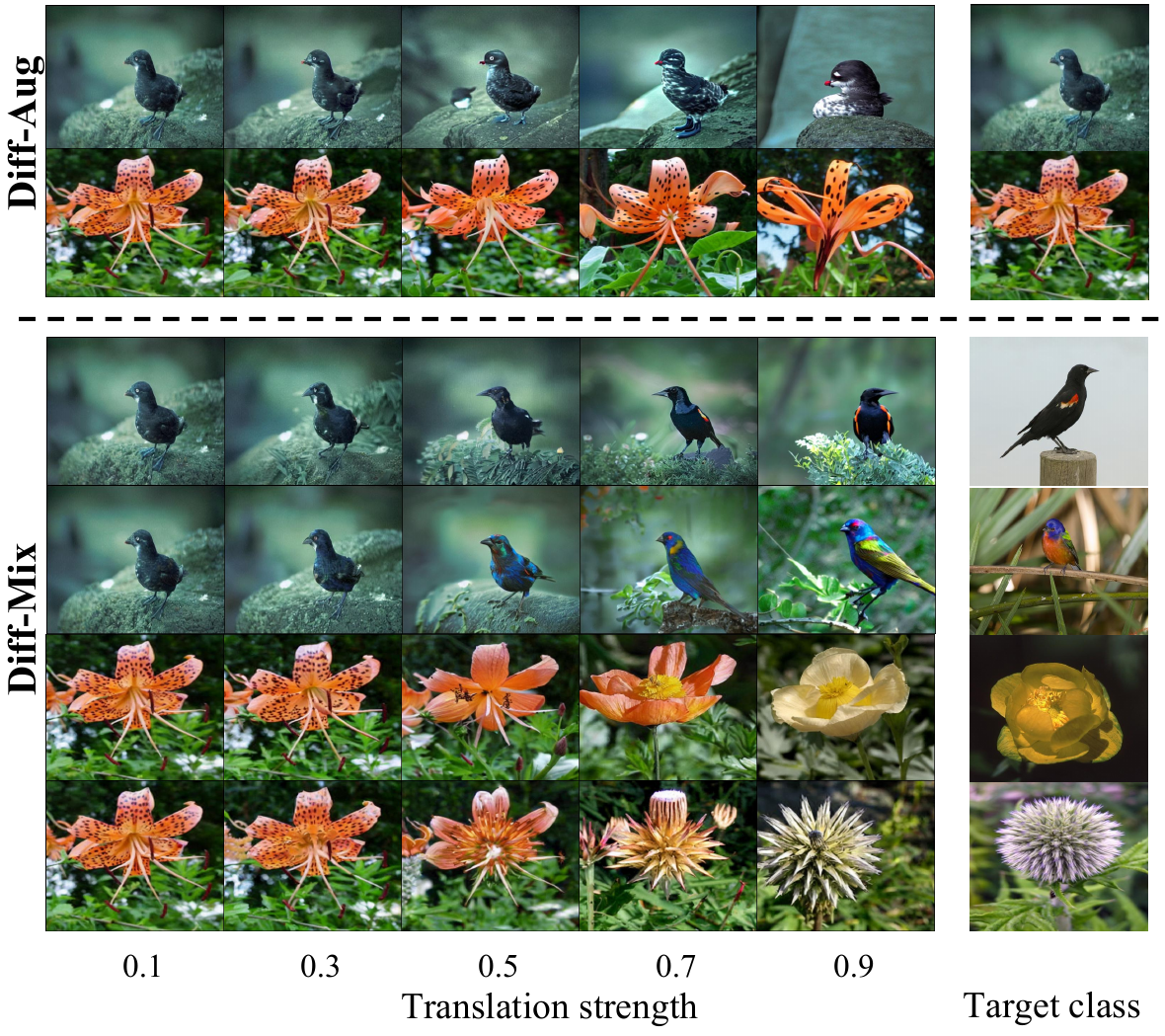}
    \caption[short]{Examples of images translated using Diff-Mix and Diff-Aug across various strengths. Diff-Aug employs the same target and reference image classes, typically resulting in subtle modifications. Diff-Mix progressively adjusts the foreground to align with the target class as the translation strength increases, while preserving the background layout from the reference image.}
  \vspace*{-0.4cm}
\end{figure}
  
\vspace{5pt}
\noindent\textbf{Diff-Mix.}
Diff-Mix employs the same translation process as Diff-Aug, but the reference image is sampled from the full training set rather than intra-class set to enable inter-class interpolation. The key difference is that Diff-Mix can generate numerous counterfactual examples, such as a blackbird in the sea (see fourth row of Fig. \ref{fig:introduction}). This necessitates that downstream models make a more refined differentiation of category attributes, thereby reducing the impact of spurious correlations introduced by variations in the background. Denoting the label of the reference image as $y^j$, by inserting the reference image into the reverse process with the prompt ``\texttt{photo of a \textup{[$\text{V}^i$] [metaclass]}}", we can obtain interpolated images between the $i_{\text{th}}$ and $j_{\text{th}}$ categories. 
By controlling the intensity $s$, we can precisely manage the interpolation process. When annotating the synthetic image, unlike Mixup and Cutmix,  we take into account the non-linear nature of diffusion translation. Thus the annotation function is given by
\begin{align}
    \tilde{y} = (1- s^\gamma) y^{i} + s^\gamma y^{j},
    \label{eq: annotation}
\end{align}
where $\gamma$ is a hyperparameter introducing non-linearity. Our empirical findings indicate that a $\gamma$ smaller than $1$ is favored. Additionally, in low-shot cases, the samples with higher confidence in the target class are preferred (see details in Sec. \ref{sec: ablation}).

\vspace{5pt}
\noindent\textbf{Construct synthetic dataset.}
To construct the synthetic dataset using Diff-Mix, similar to Da-fusion \cite{dafusion}, we adopt a randomized sampling strategy ($s\in \{0.5, 0.7, 0.9\}$) for the selection of translation strength. 
While applying the inter-class editing, we observe that Diff-Mix tends to produce more undesirable samples compared to Diff-Aug. 
These undesirable samples have incomplete foreground such as fragmented bird bodies. This is caused by the intrinsic shape and pose differences among classes.
To mitigate this, we introduce a simple data-cleaning approach to reduce the proportion of such problematic images. 
We utilize the large vision language model CLIP \cite{clip} to assess the confidence in the content, serving as the filtering criterion. 
Further details can be found in the supplementary materials (SMs). 

\begin{figure}
    \centering
    \includegraphics[width=1\linewidth]{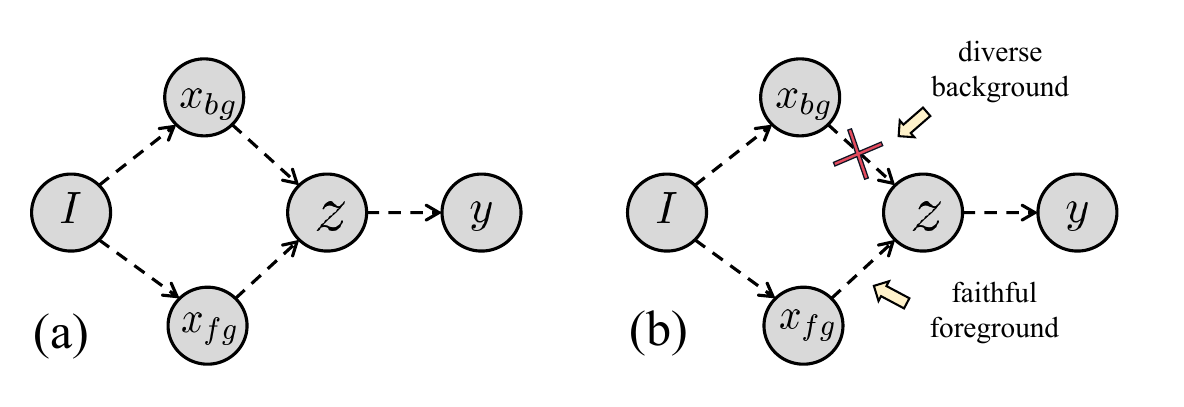}
    \caption{A schematic explanation of Diff-Mix's effectiveness using structural casual model \cite{pearl2009causal}. $\bm x_{fg}$ is the foreground that determines the real class label, $\bm x_{bg}$ denotes the background. $\bm x_{fg}\rightarrow \bm z \rightarrow y$ is the causal path that we are focusing and $\bm x_{fg} \leftarrow I \rightarrow \bm x_{bg} \rightarrow \bm z \rightarrow y$ is the backdoor path that introduces spurious relations between $\bm x_{fg}$ and $y$. }
    \label{fig: invariant_learning}
    \vspace{-0.4cm}
\end{figure}
\vspace{5pt}
\noindent\textbf{Analysis.}
We depict the core insight of Diff-Mix in Fig. \ref{fig: invariant_learning}.  To eliminate the spurious correlation introduced by $\bm x_{bg}$, learning on the synthetic set with randomized $\bm x_{bg}$ (background) can cut off spurious correlation, forcing the classification model to infer only from the foreground. The study in Fig. \ref{fig: size_diversity} (b) shows that the more diverse the background (larger the referable class number), the better the performance on the CUB test set. 





\begin{figure*}[htp]
    \begin{subfigure}{0.235\linewidth}
      \includegraphics[width=1\linewidth]{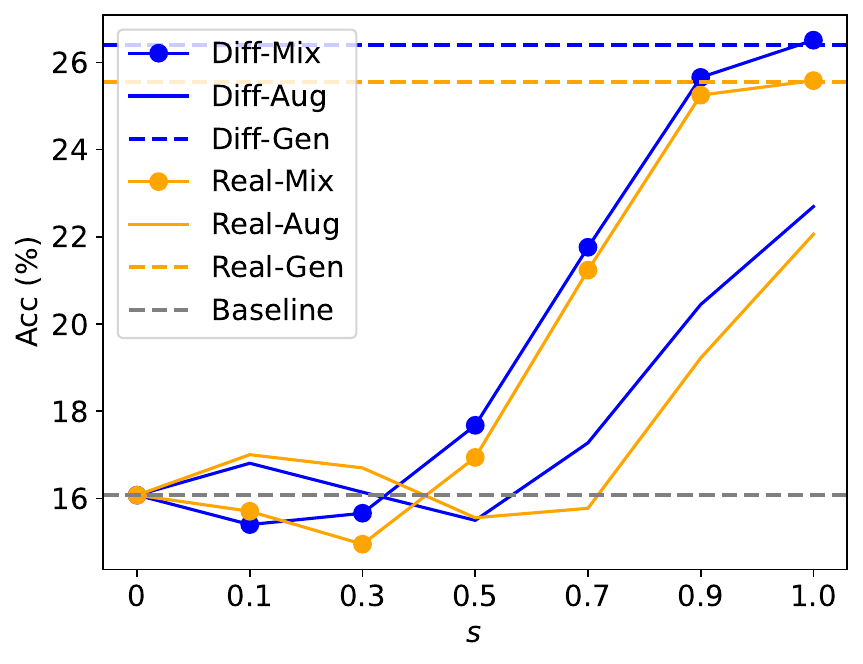}
      \caption{1-shot}
    \end{subfigure}
    \begin{subfigure}{0.235\linewidth}
      \includegraphics[width=1\linewidth]{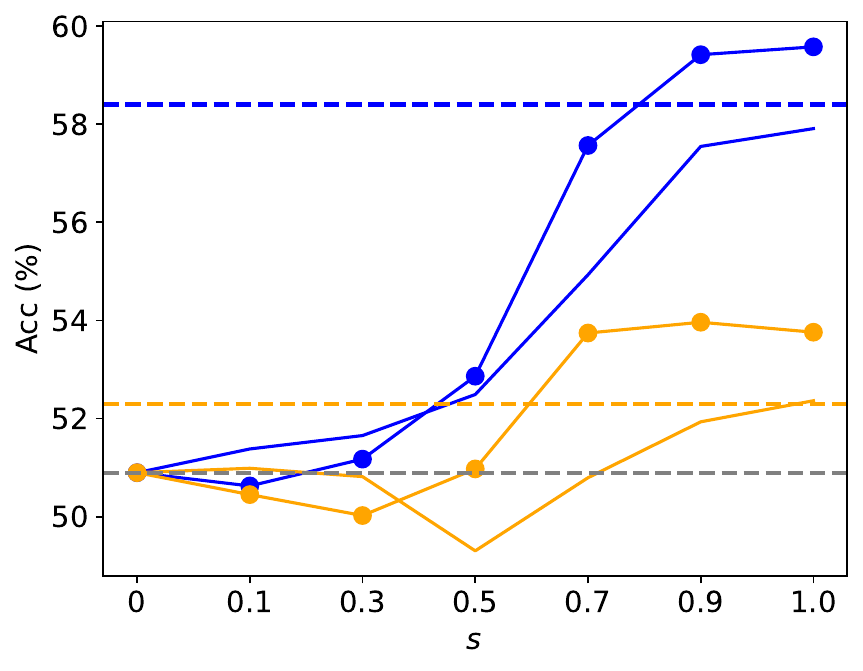}
      \caption{5-shot}
    \end{subfigure}
    \begin{subfigure}{0.235\linewidth}
        \includegraphics[width=1\linewidth]{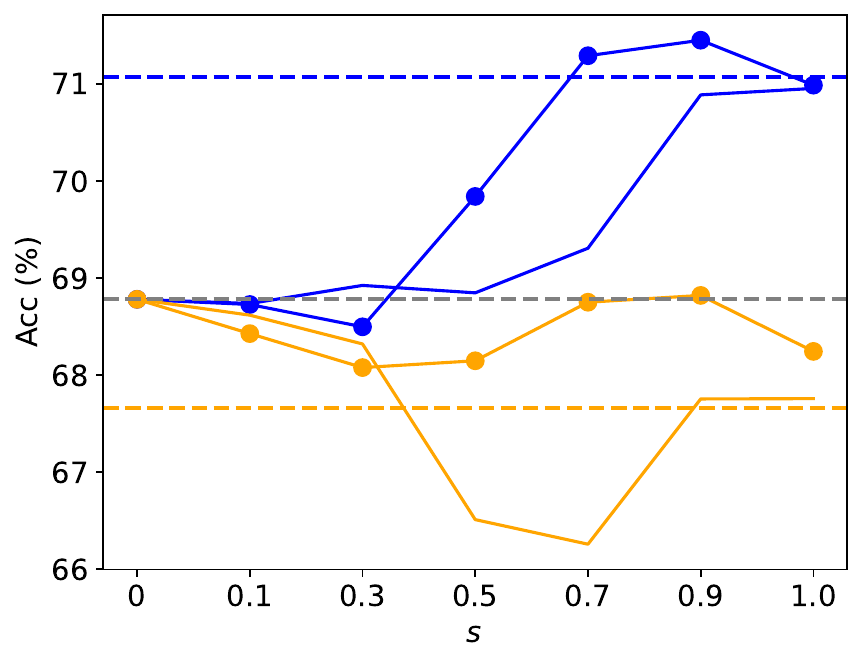}
        \caption{10-shot}
      \end{subfigure}
      \begin{subfigure}{0.235\linewidth}
        \includegraphics[width=1.03\linewidth]{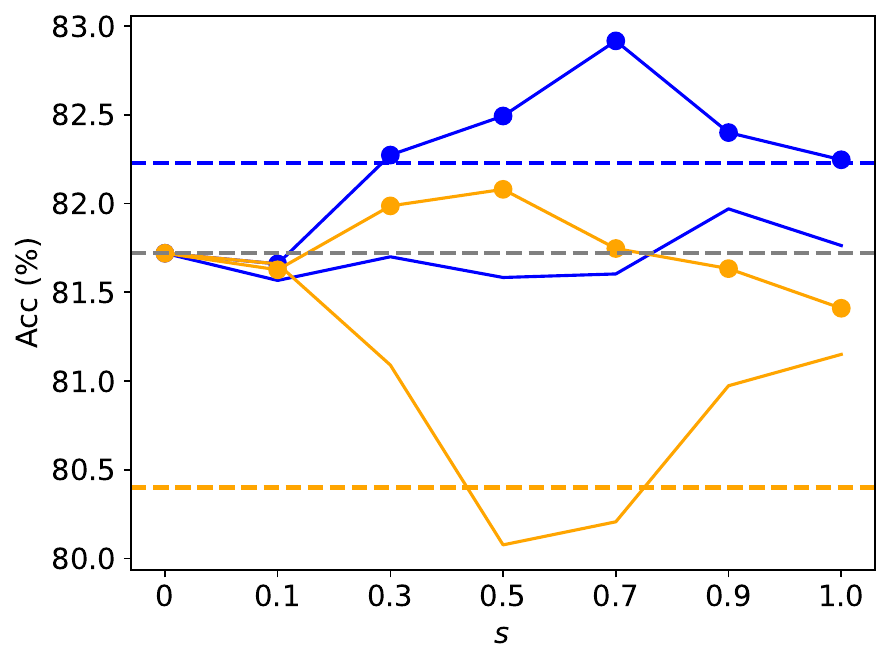}
        \caption{All-shot}
      \end{subfigure}
    

    \caption{Few-shot classification results on CUB. \label{fig: few-shot classification}} 
\vspace{-0.3cm}
    \end{figure*}
    
\section{Experiments}
In this section, we investigate the effectiveness of Diff-Mix in domain-specific datasets.
The key questions we aim to address are as follows:

\begin{itemize}
    \item[] \textbf{Q1:} Can generative inter-class augmentation lead to more significant performance gains in downstream tasks compared to those intra-class augmentation methods and distillation-based methods?
    \item[] \textbf{Q2:} Is improved background diversity the secret weapon of Diff-Mix for enhancing the performance?
    \item[] \textbf{Q3:} How to choose the fine-tuning strategy and annotation strategy to boost the performance gains for the inter-class augmentation?
\end{itemize}
To address Q1, we separately discuss these questions in few-shot settings in Sec. \ref{sec:low-shot classification}, conventional classification in Sec. \ref{sec:conventional classification}, as well as long-tail classification in Sec. \ref{sec:imbalance classification}.  
Additionally, to answer Q2, we conduct a test for background robustness in Sec. \ref{sec: robustness} and perform an ablation study focusing on the size and diversity of synthetic data in Sec. \ref{sec: ablation}.
For Q3, we conduct an ablation study to empirically discover effective strategies for deployment in Sec. \ref{sec: ablation}. 
%


\subsection{Few-shot Classification}
\label{sec:low-shot classification}
\noindent\textbf{Experimental Setting.} To investigate the impact of different data expansion methods, we conduct few-shot experiments on a domain-specific dataset Caltech-UCSD Birds (CUB) \cite{bird}, with shot numbers of 1, 5, 10, and all.  
For augmentation-based methods, the synthetic dataset is constructed using various translation strengths ($s$), specifically, $s\in\{0.1,0.3,...,1.0\}$. We expand the original training set with a multiplier of 5 and cache the synthesized dataset locally for joint training. Real data are replaced by synthetic data proportionally during training, and the replacement probability $p$ is set as 0.1, 0.2, 0.3, and 0.5 for all-shot, 10-shot, 5-shot, and 1-shot classification, respectively. All experiments use ResNet50 with an input resolution of $224 \times 224$. Additional details can be found in the SMs.

\vspace{5pt}
\noindent\textbf{Comparison methods.} 
To unveil the trade-off between faithfulness and diversity resulting from different expansion strategies, we compared \textbf{Diff-Mix} with \textbf{Diff-Gen} and \textbf{Diff-Aug}. Furthermore, we conduct experiments on expansion strategies using vanilla SD: \textbf{Real-Mix}, \textbf{Real-Gen}, and \textbf{Real-Aug}, where `Real' signifies that SD is not fine-tuned.

\vspace{5pt}
\noindent\textbf{Main results.} 
To answer Q1 under the few-shot classification setting, we augment CUB using X-Mix, X-Aug, and X-Gen, where `X' denotes `Diff/Real' for simplicity. 
The results are shown in Fig. \ref{fig: few-shot classification}, and we can observe that:
\begin{enumerate}
  \item Diff-Mix generally outperforms the intra-class competitor X-Aug and distillation competitor X-Gen in various few-shot scenarios. It tends to achieve higher gains when the strength $s$ is relatively large, \ie, \{0.5, 0.7, 0.9\}, where the foreground has been edited to match the target class and the background retains similarities to the reference image.
  \item Among the Real-X methods, distillation tends to be more effective than the augmentation method when the shot number is low, but the trend reverses as the shot number increases (compare Real-Gen with Real-Aug). Real-Gen's samples even show less effective than Real-Aug ($s=1.0$) under the all-shot case \footnote{Real-Aug ($s=1.0$) remains analogous to the reference image (slightly higher faithfulness compared to Real-Gen) because the discrete forward process cannot approximate the ideal normal distribution within a limited number of steps ($T=25$).}. This indicates that the importance of faithfulness in the trade-off between faithfulness and diversity increases with the shot number. Additionally, Real-Mix exhibits consistent and stable improvement over the other two methods.
  \item Diff-Gen consistently outperforms Real-Gen under four scenarios. Notably, Real-Gen's performance declines below that of the baseline as the shot numbers reach 10, showcasing the importance of the fine-tuning process which increases the faithfulness of synthetic samples.

\end{enumerate}


\subsection{Conventional Classification}
\label{sec:conventional classification}
\noindent\textbf{Experimental setting.} 
To test whether Diff-Mix can further boost performance in a more challenging setting, \ie, under the all-shot scenario with high input resolution, we conduct conventional classification on five domain-specific datasets: CUB \cite{bird}, Stanford Cars \cite{cars}, Oxford Flowers \cite{flower}, Stanford Dogs \cite{dog}, and FGVC Aircraft \cite{aircraft}. 
Two backbones are employed: pretrained (ImageNet1K \cite{Imagen}) ResNet50 \cite{resnet} with input resolution $448\times448$, and pretrained (ImageNet21K) ViT-B/16 \cite{vit} with input resolution $384\times384$. 
Label smoothing \cite{muller2019does} is applied across all datasets with a confidence level of 0.9. 
For all expansion strategies, the expansion multiplier is 5 and the replacement probability $p$ is 0.1.
Besides, we use a randomized sampling strategy ($s\in\{0.5,0.7,0.9\}$) and a fixed $\gamma$ (0.5, and this is specific to Diff-Mix) for all datasets.
%

\vspace{5pt}
\noindent\textbf{Comparison methods.} We compare Diff-Mix with (1) Real-Filtering (RF) \cite{SyntheticData}, a variation of Real-Gen that incorporates clip filtering, (2) Real-Guidance (RG) \cite{SyntheticData}, which augments the dataset using intra-class image translation at low strength ($s=0.1$), (3) Da-Fusion \cite{dafusion}, a method that solely fine-tunes the identifier to personalize each class and employs randomized sampling strategy ($s\in\{0.25,0.5,0.75,1.0\}$), and non-generative augmentation methods (4) CutMix \cite{yun2019cutmix} and (5) Mixup \cite{zhang2017mixup}.

\begin{table*}[thp]
    \centering
    \renewcommand{\multirowsetup}{\centering}
    \resizebox{0.88\textwidth}{!}{
    \footnotesize
        \begin{tabular}{llccccccccc}
            \hline
            \multirow{2}{*}{Backbone} & \multirow{2}{*}{Aug. Method} & \multirow{2}{*}{FT Strategy}  & \multicolumn{6}{c}{Dataset} \\
            \cline{4-10}
                                            &                                           &            & CUB           & Aircraft      & Flower        & Car                  & Dog      & Avg    &   \\\hline
            \multirow{8}{*}{ResNet50@448}   & -                                         &-           & 86.64         & 89.09         & 99.27         & 94.54           & 87.48     & 91.40   &   \\
                                            & Cutmix\cite{yun2019cutmix}                &-           & 87.23         & 89.44         & 99.25         & 94.73           & 87.59     & 91.65\textcolor{green}{$_{+0.25}$}   &   \\
                                            & Mixup\cite{zhang2017mixup}                &-           & 86.68         & 89.41         & 99.40         & 94.49           & 87.42     & 91.48\textcolor{green}{$_{+0.08}$}   &   \\\cline{2-10}
                                            & Real-filtering \cite{SyntheticData}$^\dagger$ & \XSolidBrush& 85.60    & 88.54         & 99.09         & 94.59           & 87.30     & 91.22\textcolor{red}{$_{-0.18}$}   &   \\
                                            & Real-guidance \cite{SyntheticData}$^\dagger$ &\XSolidBrush& 86.71      & 89.07         & 99.25         & 94.55           & 87.40     & 91.59\textcolor{green}{$_{+0.19}$}   &   \\
                                            & Da-fusion \cite{dafusion}$^\dagger$       & TI         & 86.30         & 87.64         &  99.37        & 94.69           & 87.33     & 91.07\textcolor{red}{$_{-0.58}$}   &   \\

                    \rowcolor{gray!40}      & Diff-Mix                                  & TI + DB    & 87.16         &\textbf{90.25} &\textbf{99.54} & 95.12           & 87.74     &91.96\textcolor{green}{$_{+0.56}$}     &   \\
                    \rowcolor{gray!40}      & Diff-Mix + Cutmix                         & TI + DB    & \textbf{87.56}& 90.01         & 99.47         & \textbf{95.21}  & \textbf{87.89}& \textbf{92.03}\textcolor{green}{$_{+0.63}$}  & \\\hline\hline
            \multirow{8}{*}{ViT-B/16@384}  & -                                         &-           & 89.37         & 83.50         & 99.56         & 94.21           & 92.06     & 91.74   &   \\
                                            & Cutmix\cite{yun2019cutmix}                &-           & \textbf{90.52}& 83.50         & 99.64         & 94.83           & \textbf{92.13}     & 92.12\textcolor{green}{$_{+0.38}$}   &   \\
                                            & Mixup\cite{zhang2017mixup}                &-           & 90.32         & 84.31         & \textbf{99.73}& 94.98           & 92.02     & 92.27\textcolor{green}{$_{+0.53}$}   &   \\\cline{2-10}
                                            & Real-filtering \cite{SyntheticData}$^\dagger$ &\XSolidBrush& 89.49     & 83.07         & 99.36         & 94.66           & 91.91     & 91.69\textcolor{red}{$_{-0.05}$}   &   \\
                                            & Real-guidance \cite{SyntheticData}$^\dagger$  &\XSolidBrush& 89.54     & 83.17         & 99.59         & 94.65           & 92.05     & 91.80\textcolor{green}{$_{+0.06}$}   &   \\
                                            & Da-fusion \cite{dafusion}$^\dagger$           & TI         &89.40      & 81.88         & 99.61         & 94.53           &92.07      & 91.50\textcolor{red}{$_{-0.24}$}   &   \\
                    \rowcolor{gray!40}      & Diff-Mix                                  & TI + DB    & 90.05         & 84.33         & 99.64         & 95.09           & 91.99     &   92.22\textcolor{green}{$_{+0.48}$} &   \\
                    \rowcolor{gray!40}      & Diff-Mix + Cutmix                         & TI + DB    & 90.35         & \textbf{85.12}& 99.68         & \textbf{95.26}  & 91.89     &   \textbf{92.46}\textcolor{green}{$_{+0.72}$} &   \\\hline

        \end{tabular}
        }
        \caption{Conventional classification in six fine-grained datasets. `$^\dagger$' indicates our reproduced results using SD.
        \vspace{-0.2cm}
        \label{ex: multi-shot classification}}
\end{table*}

\vspace{5pt}
\noindent\textbf{Main results.} 
We show the classification accuracy for different data expansion strategies in Table \ref{ex: multi-shot classification}, our observations can be summarized: (1) Diff-Mix consistently demonstrates stable improvements across the majority of settings. Its average performance gain across the five datasets exceeds that of baselines employing intra-class augmentation methods (RG and Da-fusion), distillation method (RF), and non-generative data augmentation techniques (CutMix and Mixup).
(2) Real-filtering, analogous to the discussion of Real-Gen, exhibits performance degradation on most datasets due to the distribution gap. 
%
(3) The combined use of Diff-Mix and CutMix often yields better performance gains. This is attributed to the distinct enhancement mechanisms of the two methods, \ie, vicinal risk minimization \cite{chapelle2000vicinal, zhang2017mixup} and foreground-background disentanglement.
(4) Diff-Mix does not exhibit significant performance improvement in the dog dataset. We attribute this lack of improvement to the complexity of the dog dataset, which often contains multiple subjects in a single image, impeding effective foreground editing (refer to the SMs for visual examples).

\subsection{Long-Tail Classification }
\label{sec:imbalance classification}

\vspace{5pt}
\noindent\textbf{Experiment setting.} Following the settings of previous long-tail dataset constructions \cite{cmo, ImageNet-LT, cao2019learning}, we create two domain-specific long-tail datasets, CUB-LT \cite{CUB-LT} and Flower-LT. The imbalance factor controls the exponential distribution of the imbalanced data, where a larger value indicates a more imbalanced distribution. To leverage generative models for long-tail classification, we adopt the approach of SYNAuG \cite{SynAug}, which uniformize the imbalanced real data distribution using synthetic data. Translation strength $s$ (0.7) and $\gamma$ (0.5) are fixed for both Diff-Mix and Real-Mix.

\begin{figure}[th]
\centering
    \begin{minipage}{0.5\textwidth}
       \resizebox{0.95\linewidth}{!}{
        \begin{tabular}{lcccccc}
        \hline
        \multirow{2}{*}{Method}        & \multicolumn{4}{c}{IF=100}       &  \multirow{2}{*}{50}     &  \multirow{2}{*}{10}\\\cline{2-5} 
                            & Many              & Medium            & Few            & All             &                &           \\\hline
        CE                  &  79.11            &   64.28           &   13.48        &  33.65          & 44.82          &   58.13   \\
        CMO \cite{cmo}            &   78.32           &   58.57           &   14.78        &  32.94          &  44.08         &   57.62   \\
        CMO + DRW \cite{drw}           &   78.97           &   56.36           &   14.66                &  32.57                    &   46.43        &   59.25   \\\hline
        Real-Gen            &  \textbf{84.88}            &  65.23            &   30.68        &  45.86          &    53.43           &   61.42   \\
        \rowcolor{gray!40}Real-Mix (s=0.7)    &  84.63            &  66.34                  &  34.44            &    47.75           &   55.67        &   62.27   \\
        \rowcolor{gray!40}Diff-Mix (s=0.7)    &  84.07            &\textbf{67.79}           &  \textbf{36.55}   &    \textbf{50.35}          &  \textbf{58.19}&   \textbf{64.48}   \\
        \hline
        \end{tabular}
        }
        \captionof{table}{Long-tail classification in CUB-LT.\label{table:cub_LT}}
    \end{minipage}
    \begin{minipage}{0.5\textwidth}
        \vspace{0.3cm}
        \resizebox{0.95\linewidth}{!}{
        \begin{tabular}{lcccccc}
        \hline
        \multirow{2}{*}{Method}                  & \multicolumn{4}{c}{IF=100}       &  \multirow{2}{*}{50}     &  \multirow{2}{*}{10}\\\cline{2-5}  
                            & Many               & Medium          & Few               & All                &               &           \\\hline
        CE                  & 99.19              &   94.95         &   58.18           &  80.43             &   90.87       &  95.07    \\
        CMO \cite{cmo}            & 99.25              &   95.19         &   67.45           &  83.95             &   91.43       &  95.19    \\
        CMO+ DRW \cite{drw}           & \textbf{99.97}              &   95.06         &   67.31           &  84.07             &   92.06       &  95.92    \\\hline
        Real-Gen            & 98.64              &   95.55         &   66.10           &  83.56             &   91.84       &  95.22    \\
        \rowcolor{gray!50}Real-Mix (s=0.7)    & 99.87     &   96.26         &   68.53           &  85.19             &   92.96       &  96.04    \\
        \rowcolor{gray!50}Diff-Mix (s=0.7)    & 99.25              & \textbf{96.98}  &\textbf{78.41}     &  \textbf{89.46}    & \textbf{93.86}&  \textbf{96.63}\\
        \hline
        \end{tabular}
        }
        \captionof{table}{Long-tail classification in Flower-LT.\label{table:flower_LT} }
 \vspace{-0.5cm}
    \end{minipage}
    \end{figure}

\vspace{5pt}
\noindent\textbf{Comparison methods.} We compare Diff-Mix with Real-Mix, Real-Gen, the non-generative CutMix-based oversampling approach CMO \cite{cmo}, and its enhanced variant with two-stage deferred re-weighting \cite{cao2019learning} (CMO+DRW). 

\vspace{5pt}
\noindent\textbf{Main results.} 
We present the long-tail classification results for CUB-LT in Table \ref{table:cub_LT} and Flower-LT in \ref{table:flower_LT}. The observations are as follows:
(1) Generative approaches exhibit superior performance in tackling imbalanced classification issues compared to CutMix-based methods (CMO and CMO+DRW).
(2) Real-Mix surpasses Real-Gen in performance across various imbalance factors in two datasets. This indicates that tail classes can benefit from the enhanced diversity by leveraging the visual context of majority classes.
(3) Diff-Mix generally achieves the best performance among the compared strategies, especially at the low-shot case, highlighting the importance of fine-tuning.

\subsection{Background Robustness}
\label{sec: robustness}
\begin{figure}
\begin{minipage}{0.475\textwidth}
  \centering
  \resizebox{0.95\linewidth}{!}{%
    \begin{tabular}{c|ccccc}
  \toprule
   Group           & Base.    & CutMix             & DA-fusion & Diff-Aug & \textbf{Diff-Mix}           \\ 
  \hline\hline
  (waterbird, water) &59.50     &62.46             & 60.90    & 61.83    &\textbf{63.83}                       \\ 
  \hline
  (waterbird, land)  &56.70     &60.12             &58.10     & 60.12    &\textbf{63.24}                       \\ 
  \hline
  (landbird, land)   &73.48     & 73.39            &72.94     & 73.04    &\textbf{75.64}                       \\ 
  \hline
  (landbird, water)  &73.97     & \textbf{74.72}   &72.77     & 73.52    &74.36                        \\ 
  \hline
  Avg.             & 70.19    & 71.23            & 69.90    &70.28     &\textbf{72.47}                             \\
  \toprule
  \end{tabular}
  }
  \captionof{table}{Classification results across four groups in Waterbird \cite{sagawa2019distributionally}. Waterbird is an out-of-distribution dataset for CUB, crafted by segmenting CUB's foregrounds and paste them into the scene images from Places \cite{zhou2017places}. The constructed dataset can be divided into four groups based on the composition of foregrounds (waterbird and landbird) and backgrounds (water and land) \label{tab: robustness}. \label{tab:robustness}}
  \vspace{-0.2cm}
\end{minipage}
\end{figure}
    
In this section, we aim to evaluate Diff-Mix's robustness to background shifts, specifically, whether synthesizing more diverse samples can improve the classification model's generalizability when the background is altered. To achieve this, we utilize an out-of-distribution test set for CUB, namely Waterbird \cite{sagawa2019distributionally}. We then perform inference on the whole Waterbird set using classifiers that have been trained on either the original CUB dataset or its expanded variations. In Table \ref{tab: robustness}, we present the classification accuracies across the four groups and compare Diff-Mix with other intra-class methods (Da-fusion and Diff-Aug) as well as CutMix. We observe that Diff-Mix generally outperforms its counterparts and achieves a significant performance improvement (+6.5\%) in the challenging counterfactual group (waterbirds with land backgrounds). It is important to highlight that the background scenes in the Waterbird dataset are novel to CUB, requiring the classification model to have a stronger perceptual capability for the images' foregrounds.

\subsection{Discussion}
In this section, we address Q2 by examining the impact of size and diversity on synthetic data. Furthermore, we perform an ablation study to assess the effects of fine-tuning strategies and training hyperparameters of Diff-Mix, which is aimed at answering Q3. Unless specified otherwise, our discussions are based on experiments conducted using CUB with ResNet50, where inputs are resized to $224 \times 224$.
\label{sec: ablation}
\vspace{5pt}

\begin{figure}
    \begin{subfigure}{0.47\linewidth}
        \centering
        \includegraphics[width=1\linewidth]{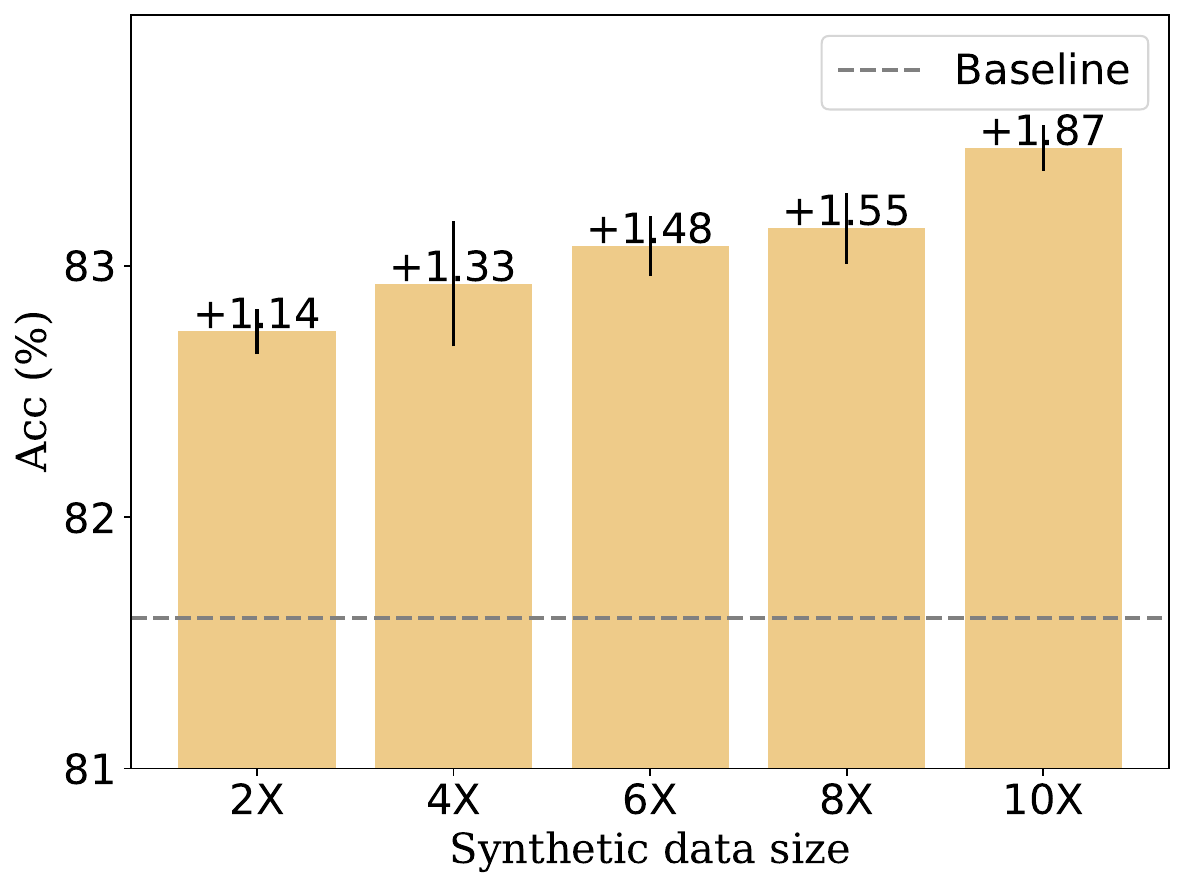}
        \caption{}
    \end{subfigure}
    \hfill
    \begin{subfigure}{0.47\linewidth}
        \centering
        \includegraphics[width=1\linewidth]{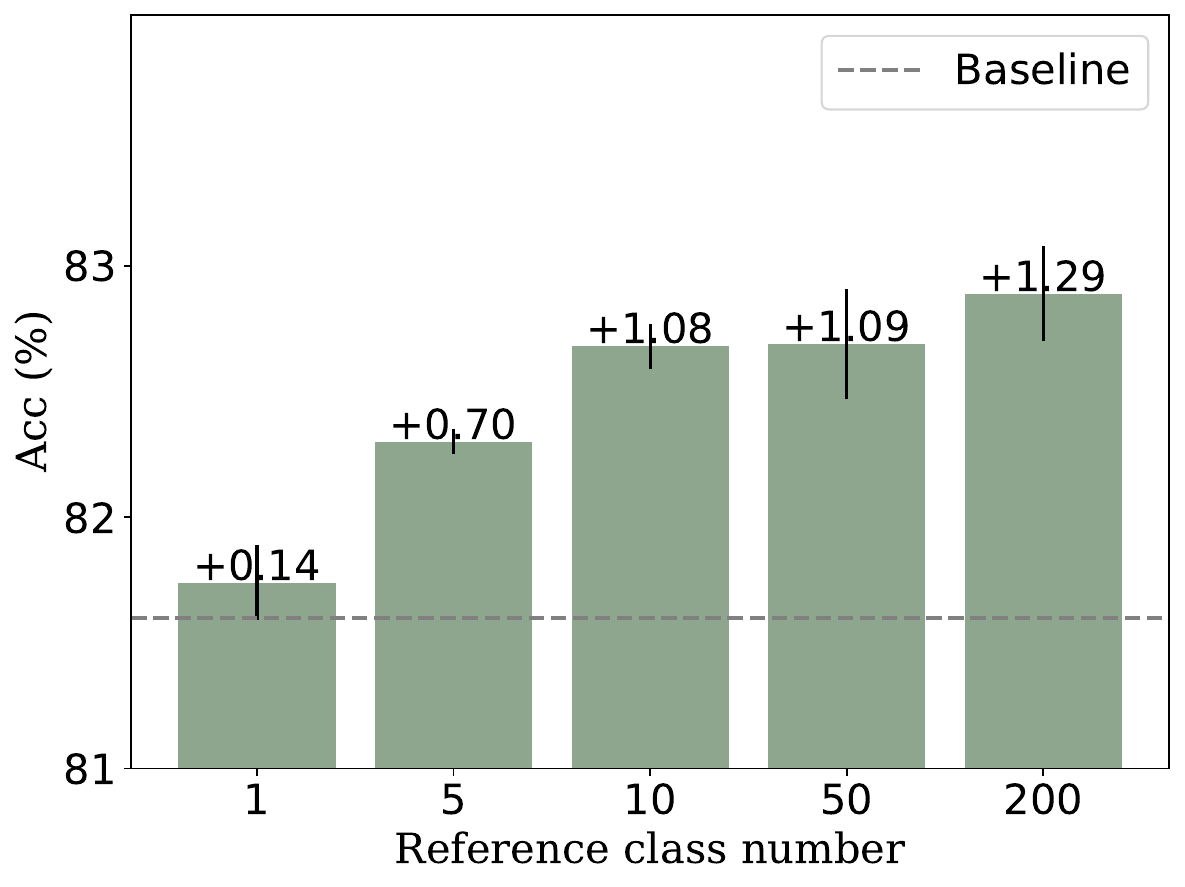}
        \caption{}
    \end{subfigure}
\caption{Comparison of results across various (a) synthetic data sizes and (b) numbers of referable classes for each target class.}
\vspace{-0.2cm}
\label{fig: size_diversity}
\end{figure}

\noindent\textbf{Size and diversity of synthetic Data.} The relationship between performance gain and the size of synthetic data is depicted in Fig. \ref{fig: size_diversity} (a), where a classification model was trained with synthetic data of varying sizes. We observe a monotonically increasing trend as the multiplier for the synthetic dataset ranges from 2 to 10. Ideally, the combination choices of $(\bm x_i, y_j)$ are in the order of $N|D^{train}|$ ($N=200$ for CUB). Furthermore, we limit the number of referable classes for each target class, which means the number of referable backgrounds decreases, resulting in a synthetic dataset of relatively lower diversity. The results are shown in Fig. \ref{fig: size_diversity} (b), and we observe a consistent improvement in performance as the number of referable classes increases. These results consistently underscore the critical role of background diversity introduced by Diff-Mix.

\vspace{5pt}
\noindent\textbf{Impact of fine-tuning strategy.} Here we compare three different fine-tuning strategies: TI, DB, and the combined TI+DB. All strategies share the same fine-tuning hyperparameters and training steps (35000). To evaluate the distribution gap, we compute the FID score \cite{fid} of synthesized images (Diff-Gen) with the training set. As illustrated in Table \ref{table: ab_finetune_strategy}, we observe that both TI+DB and TI have lower FID scores than DB. This can be attributed to the fact that semantic proximity impedes the convergence process. Additionally, while using TI alone results in a relatively low FID score, the improvements in performance are limited. This limitation stems from TI's inability to accurately reconstruct detailed concept (foreground) information, as it is primarily fine-tuned at the semantic level \cite{customdiffusion}.


\vspace{5pt}
\noindent\textbf{Annotation function.} 
In this section, we discuss the impact of the choices of the translation strength $s$ and the non-linear factor $\gamma$ in Eq. \ref{eq: annotation}. As shown in Table \ref{table: ab_gamma}, we observe that as the translation strength decreases, the optimal value for $\gamma$ also decreases, which underscores the non-linearity of Diff-Mix. The comparison between the 5-shot and all-shot settings indicates that the model tends to prefer a more diverse synthetic dataset when the number of training shots is small ($s = 0.9$ for 5-shot, $s=0.7$ for all-shot). Besides, a larger confidence in the target class is preferred when the shot number is small ($\gamma = 0.1$ for 5-shot, $\gamma=0.5$ for all-shot). A possible explanation is that the all-shot setting is less tolerant towards unrealistic images, as discussed in Section \ref{fig: few-shot classification}. Empirically, we recommend choosing a higher translation strength ($s \in {0.5, 0.7, 0.9}$) and a smaller $\gamma$ ($\gamma \in {0.1, 0.3, 0.5}$) as a conservative option.

    \begin{figure}
        \begin{minipage}{0.475\textwidth}
            \footnotesize
            \centering
            \resizebox{0.95\linewidth}{!}{
            \begin{tabular}{llcccc}
            \hline
                  & & Baseline &  TI     &    DB & TI + DB    \\\hline  
           \multirow{2}{*}{5-shot}  &FID (Diff-Gen)          &  -     & \textbf{18.26} & 19.55   &18.43  \\
            &Acc. (Diff-Mix)        &   50.90   &  57.64   & 56.11  & \textbf{59.41} \\\hline
           \multirow{2}{*}{All-shot} &FID (Diff-Gen)        & -        & 14.13 & 14.64 &\textbf{13.99}  \\
            &Acc. (Diff-Mix)    &  81.60   &    81.86  &  81.99 & \textbf{82.85} \\\hline
            \end{tabular}
            }
            \captionof{table}{Comparison of distribution gap and classification accuracy across three fine-tuning strategies. TI solely fine-tunes the identifier, and DB solely fine-tunes the U-Net, and TI+DB. \label{table: ab_finetune_strategy}}
        \end{minipage}
    \end{figure}

\begin{figure}
    \begin{minipage}{0.475\textwidth}
    \centering
    \footnotesize
    \resizebox{0.95\linewidth}{!}{
    \begin{tabular}{lcccccc}
        \toprule
        \multirow{2}{*}{$\gamma$} &\multicolumn{3}{c}{5-shot}&\multicolumn{3}{c}{All-shot}                                               \\\cmidrule(lr){2-4}
        \cmidrule(lr){5-7}
              &      $s=0.5$     &      $s=0.7$         & $s=0.9$            &      $s=0.5$                  &      $s=0.7$              & $s=0.9$                     \\\hline
        1.5   &    -4.50        &     -0.31            &   +10.30              &       -1.08                 &     +0.92              & \textbf{+0.90}           \\
        1.0   &    -2.31        &       +2.99            &   +10.79             &      +0.25                 &     +1.14              & \textbf{+0.90}           \\
        0.5   &    +2.35         &      +8.44            &   +11.01             &      +0.92                 &     \textbf{+1.30}     & +0.86                    \\
        0.3   &    +3.94         &      +9.41            &   \textbf{+11.15}    &      \textbf{+0.97}        &     +1.24              & +0.69                    \\
        0.1   &    \textbf{+6.18}&      \textbf{+9.86}   &   +10.84             &      +0.50                 &     +0.88              & +0.84                    \\
        0.0   &    +5.25         &      +9.41            &   +11.06             &      +0.38                 &     +0.63              & +0.54                    \\\bottomrule
        \end{tabular}
    }

    \captionof{table}{Comparison of performance gain across various $\gamma$ and translation strength $s$. Lower  $\gamma$ indicates a higher confidence over target class, \eg $(\gamma=0.1,s=0.7)$ results in $0.04y_i + 0.96y_j$ and $(\gamma=0.5,s=0.7)$ results in $0.16y_i + 0.84y_j$. \label{table: ab_gamma}}
    \end{minipage}

\end{figure}




\section{Conclusion}
In this work, we investigate two pivotal aspects, faithfulness and diversity, that are critical for the current state-of-the-art text-to-image generative models to enhance image classification tasks. To achieve a more effective balance between these two aspects, we propose an inter-class augmentation strategy that leverages Stable Diffusion. This method enables generative models to produce a greater diversity of samples by editing images from other classes and shows consistent performance improvement across various classification tasks.

\section{Acknowledgement}
This research is mainly supported by the National Natural Science Foundation of China (92270114). This work is also partially supported by the National Key R\&D
Program of China under Grant 2021ZD0112801.



\clearpage
{
    \small
    \bibliographystyle{ieeenat_fullname}
    \bibliography{main}
}
\clearpage
\setcounter{page}{1}
\maketitlesupplementary
\section{Appendix}
\
This appendix is organized as follows:
\begin{itemize}
    \item In Sec. \ref{sec:appendix_dc}, we elaborate on the details of data cleaning design for Diff-Mix.
    \item In Sec. \ref{sec:appendix_visualization}, additional visualizations are presented, including the visualization of attention maps and failure examples of complex datasets.
    \item  In Sec. \ref{sec:append_experiment}, few-shot classification results on a general dataset Pascal is provided \cite{pascal}.
    \item  In Sec. \ref{sec:appendix_experiment_details} and Sec. \ref{sec:appendix_latency}, implementation details and latency considerations are presented, respectively.
\end{itemize}

\subsection{Data-Cleaning Strategy}
\begin{figure}
    \centering
    \includegraphics[width=1\linewidth]{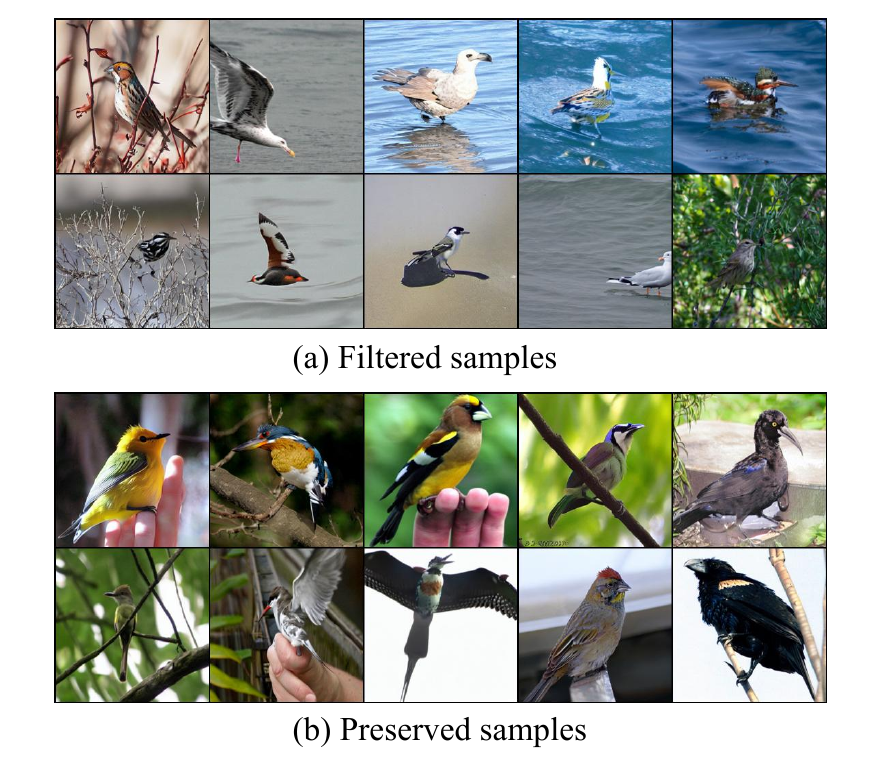}
    \caption{Examples of (a) filtered samples with the lowest 10\% confidence scores and (b) preserved samples in the CUB dataset.}
    \label{fig:appendix_filtered_samples}
\end{figure}
\label{sec:appendix_dc}
Due to the inherent differences in contour and size between the two classes, there is a higher risk of producing less realistic images during inter-class editing. We employ a simple data cleaning strategy that utilizes CLIP \cite{clip}\footnote{\href{https://huggingface.co/openai/clip-vit-base-patch32}{https://huggingface.co/openai/clip-vit-base-patch32}} as the filtering criterion. Specifically, we construct a positive caption, "a photo with a [metaclass] on it", and a negative caption, "a photo without a [metaclass] on it", and evaluate the synthetic data's confidence score towards the positive caption. We filter out the 10\% of samples with the lowest confidence scores (we do not synthesize an additional 10\% samples after data cleaning), and a subset of the filtered samples is displayed in Fig. \ref{fig:appendix_filtered_samples}. The preserved samples constitute the synthetic dataset that participates in the training process of downstream classification tasks.



\subsection{Visualizations}
\label{sec:appendix_visualization}

\begin{figure*}
    \centering
    \includegraphics[width=1\linewidth]{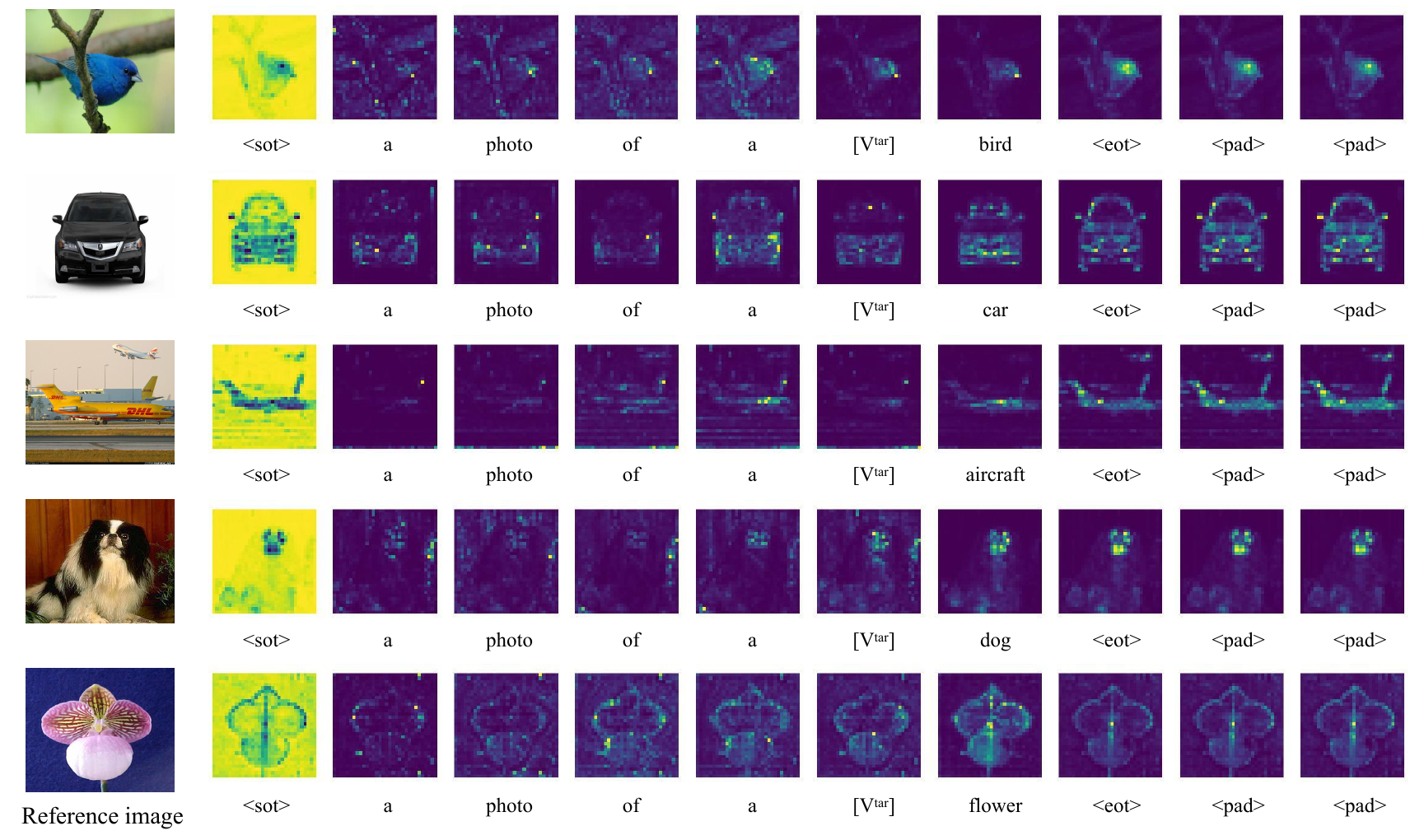}
    \caption{Visualizations of attention maps are shown in different rows for various datasets: CUB (Row 1) \cite{bird}, Stanford Cars (Row 2) \cite{cars}, FGVC Aircraft (Row 3) \cite{aircraft}, Stanford Dogs (Row 4) \cite{dog}, and Oxford Flowers (Row 5) \cite{flower}. These attention maps were generated during inter-class editing using Diff-Mix.}
    \label{fig:appendix_attn_map}
\end{figure*}

\noindent\textbf{Visualizations of attention maps.} 
In Section \ref{sec:method_data_synthesis}, we have shown that Diff-Mix can perform foreground editing while preserving most of the layout of the reference image. To support the claim, we provide evidence that SD can offer weak segmentation through textual conditions and achieve realistic foreground editing. We present visualizations of attention maps in Figure \ref{fig:appendix_attn_map} for different datasets.
 The identifier, class descriptor (e.g., ``\texttt{bird}'', ``\texttt{car}'') and the ``\texttt{<eot>}'' token, which contains the global semantic information, tend to attend to the foreground part in the reference image. For example, the mentioned tokens primarily emphasize the bird rather than the tree branches (refer to Row 1 in the figure). This suggests that textual guidance, can offer a robust foreground prior, aiding in effective foreground editing at intermediate strengths. We posit that this characteristic ensures the generation of challenging samples where the foreground is replaced by the target concept when employing Diff-Mix for inter-class editing.

\begin{figure*}
\begin{minipage}{0.95\linewidth}
    \centering
    \includegraphics[width=0.85\linewidth]{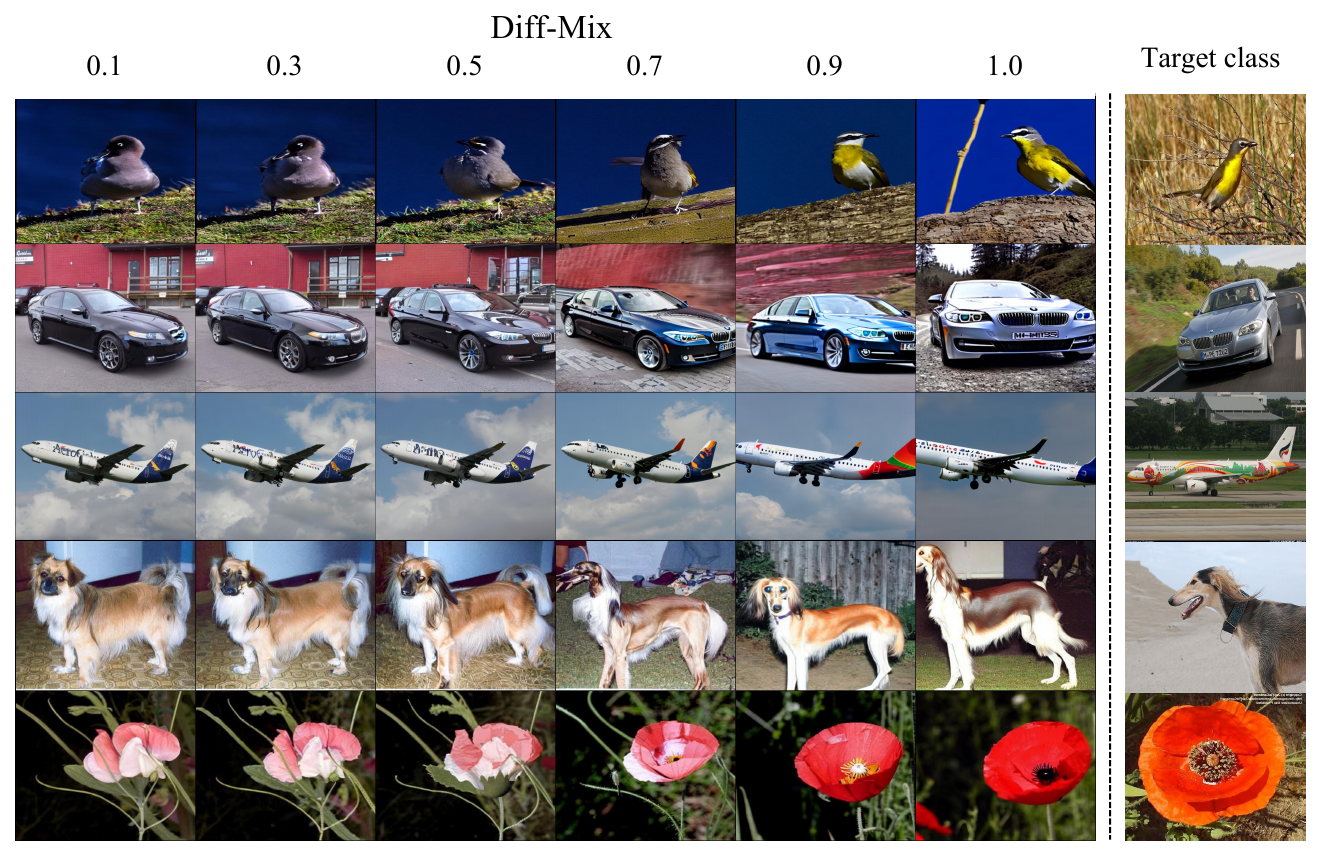}
    \caption{Examples of image generated using Diff-Mix with varying translation strengths in CUB (Row 1), Stanford Cars (Row 2), FGVC Aircraft (Row 3), Stanford Dogs (Row 4) , Oxford Flowers (Row 5). }
    \label{fig:appendix_more_visulizations}
\end{minipage}
\begin{minipage}{1\linewidth}
\centering
\vspace{0.5cm}
\includegraphics[width=0.87\linewidth]{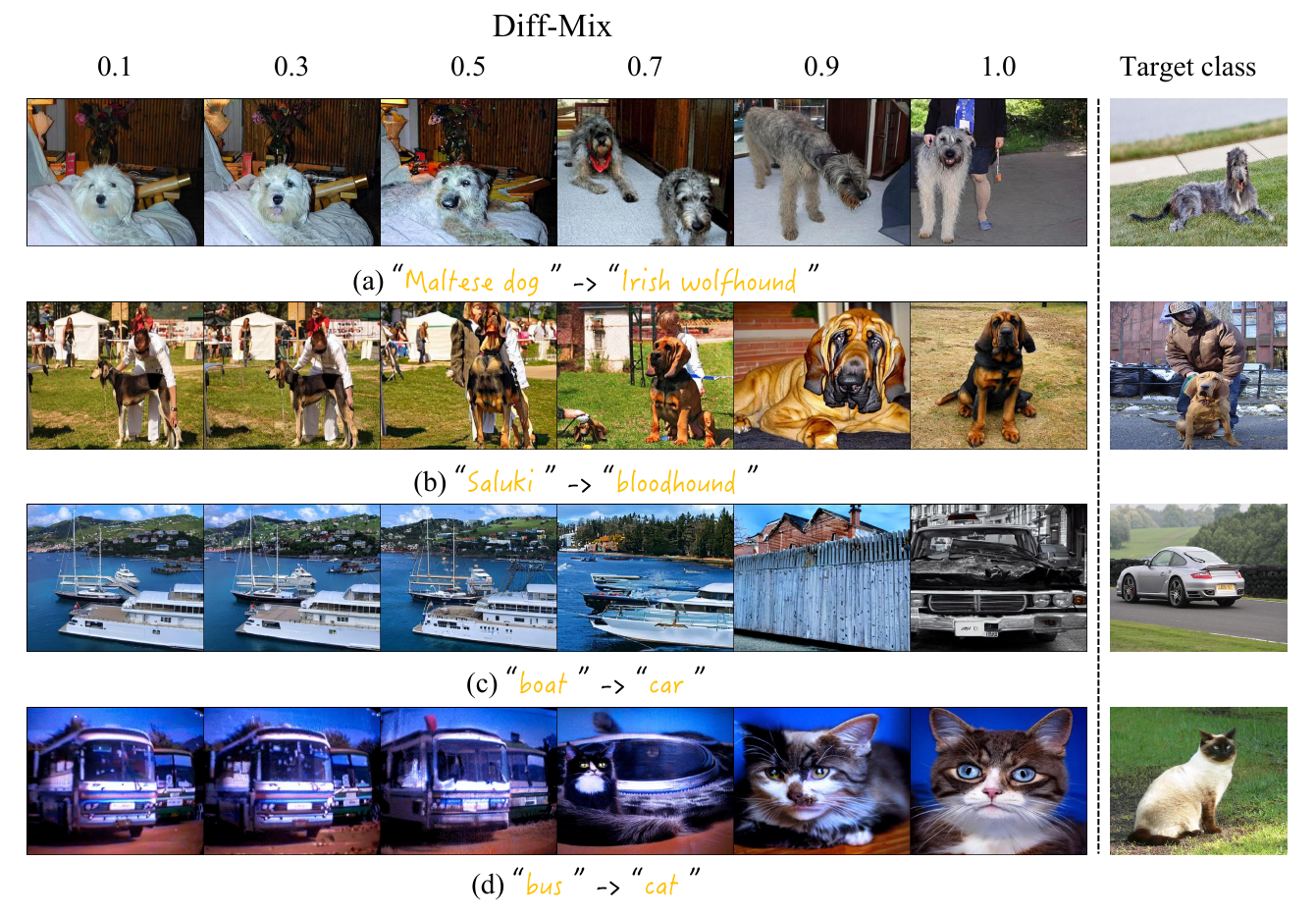}
    \caption{Failure examples generated using Diff-Mix with varying translation strengths are shown in panels (a) and (b) for the complex dataset Stanford Dogs, and in panels (c) and (d) for the general dataset Pascal \cite{pascal}.}
    \label{fig:appendix_failure_cases}
\end{minipage}

\end{figure*}

\vspace{5pt}
\noindent\textbf{Visualizations on more datasets.} In Fig. \ref{fig:appendix_more_visulizations}, we illustrate the editing process of Diff-Mix with varying strength $s\in\{0.1,0.3,0.5,0.7,0.9,1.0\}$ across five datasets. It is worth noting that for strength $s=1$, the translated images still exhibit a certain degree of similarity to the reference images. This phenomenon may be attributed to the last time-step not guaranteeing a zero signal-to-noise ratio, preserving the style and layout of the reference images \cite{lin2023common}. Particularly, when the foreground is distinct against a simple background, Diff-Mix tends to generate high-quality interpolated images. We also observe that for more complex datasets, such as Stanford Dogs, where the foreground is less clear, and there are multiple concepts in a single image, unrealistic images tend to be generated, as seen in Fig. \ref{fig:appendix_failure_cases} (a) and (b). For general dataset Pascal \cite{pascal}, the dramatic differences in contour and size between two distinct classes lead to the generation of more unrealistic images (\eg, ``\texttt{bus}" $\longrightarrow$ ``\texttt{cat}"), especially at intermediate strengths (\eg, 0.7), as seen in Fig \ref{fig:appendix_failure_cases} (c) and (d).



\begin{figure*}
\begin{minipage}{1\linewidth}
    \centering
    \includegraphics[width=1.04\linewidth]{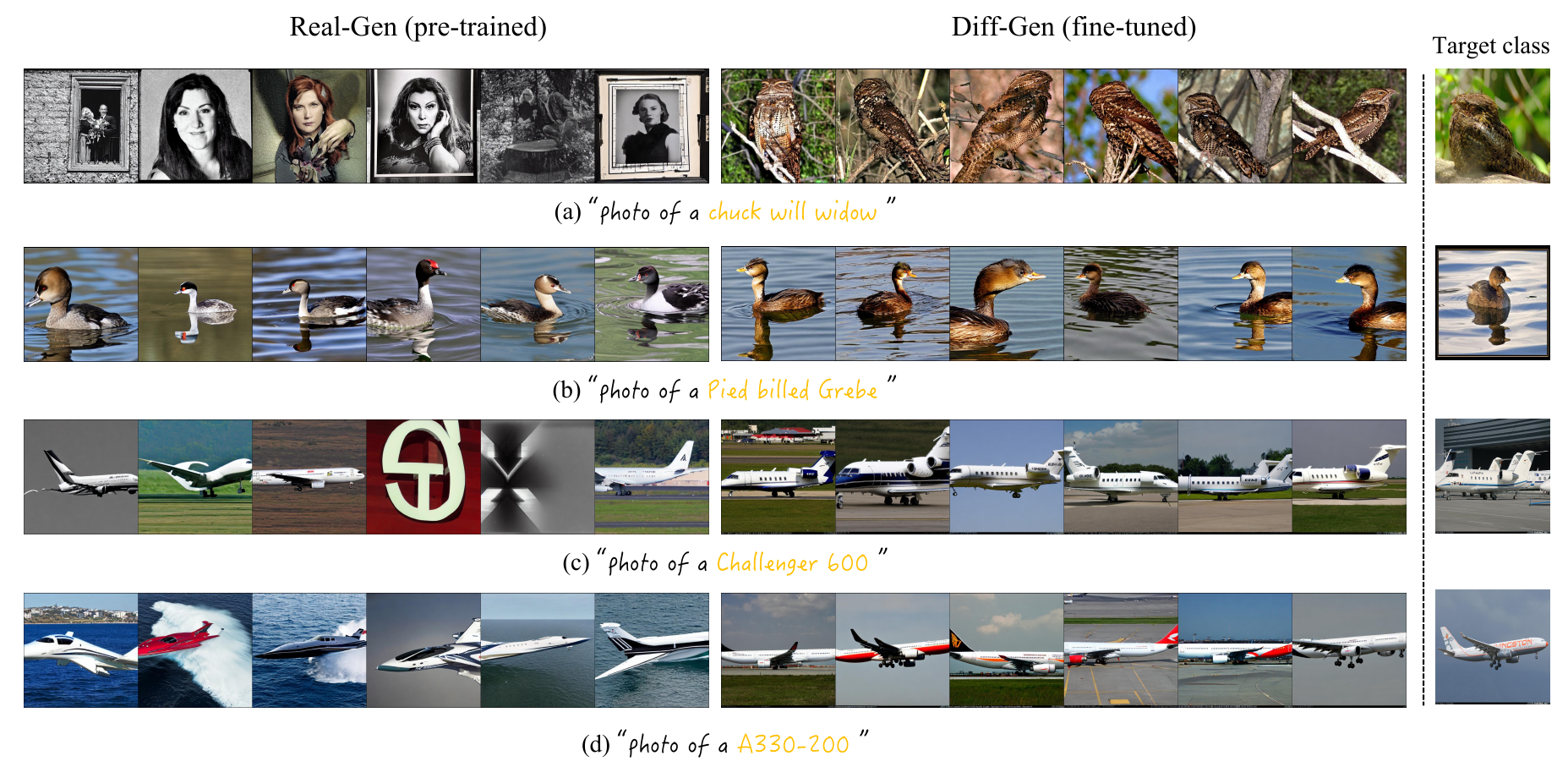}
    \caption{Examples of image generated using Real-Gen and Diff-Gen. The prompts are formatted as ``\texttt{photo of a [terminology name]}'' for Real-Gen and ``\texttt{photo of a [V$^i$] [metaclass]}'' for Diff-Gen. Panels (a) and (b) depict the samples of CUB dataset, while panels (c) and (d) depict the samples of FGVC Aircraft dataset.  }
    \label{fig:appendix_realgen_diffgen}
\end{minipage}

\begin{minipage}{1\linewidth}
\vspace{1.5cm}
\centering
\includegraphics[width=1.04\linewidth]{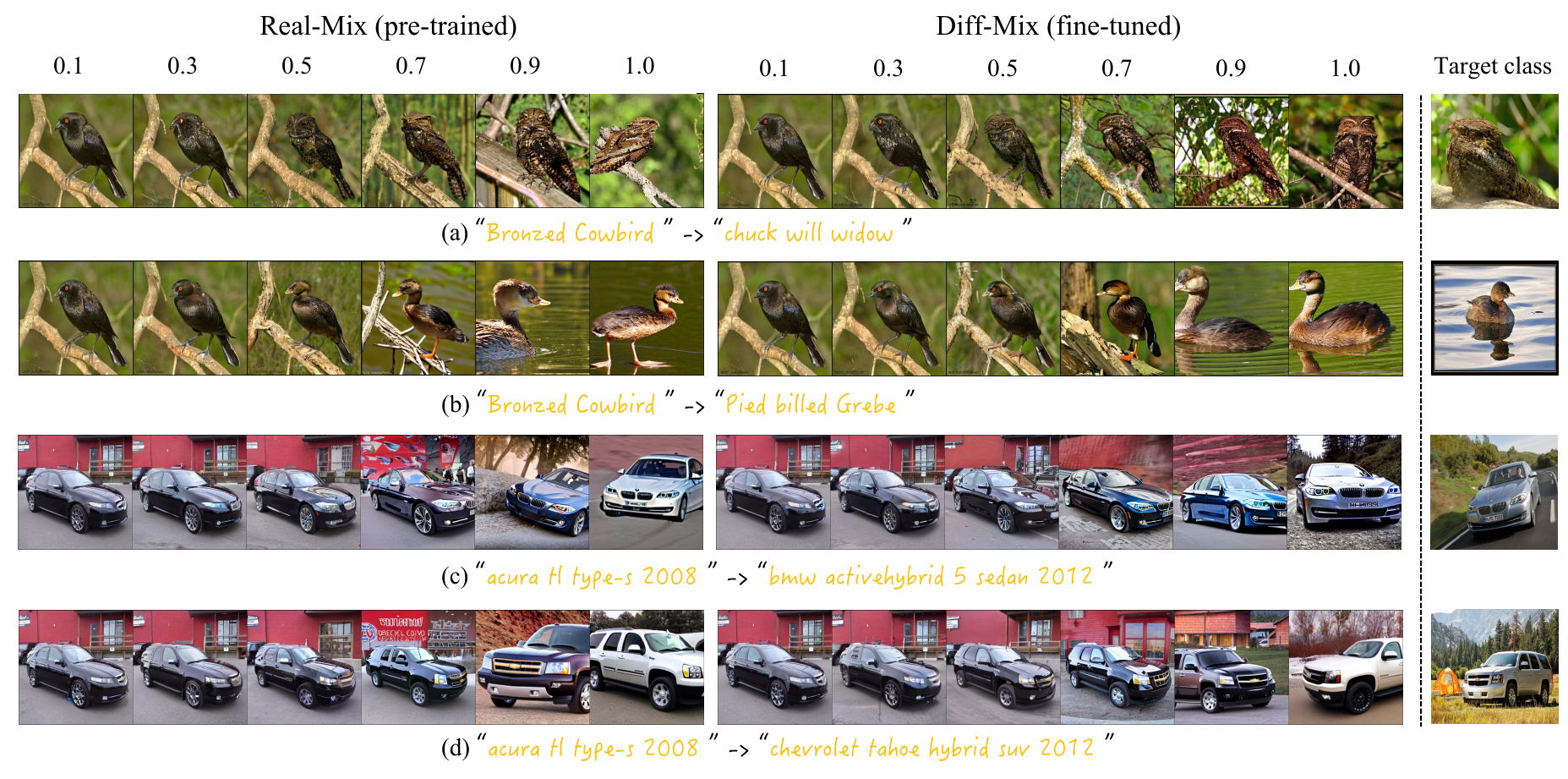}
    \caption{Examples of image generated using Real-Mix and Diff-Mix with varying translation strengths. The prompts are formatted as ``\texttt{photo of a [terminology name]}'' for Real-Mix and ``\texttt{photo of a [V$^i$] [metaclass]}'' for Diff-Mix. Panels (a) and (b) depict the samples of CUB dataset, while panels (c) and (d) depict the samples of Stanford Car dataset.  }
    \label{fig:appendix_realmix_diffmix}
\end{minipage}

\end{figure*}

\vspace{5pt}
\noindent\textbf{Real-Gen versus Diff-Gen.} To illustrate the distribution gap between domain-specific datasets and the pre-trained T2I model, as well as to demonstrate how fine-tuning can significantly mitigate this gap, we present a comparison in Fig. \ref{fig:appendix_realgen_diffgen}. It is worth noting that Real-Gen sometimes fails to generate correct concepts based on the terminology name of the target class (see ``\texttt{photo of a chuck will widow}" in panel (a)). Diff-Gen tends to generate more faithful outputs, it is noted that the majority of the generated images exhibit a similar layout and closely resemble the training samples. This resemblance is especially pronounced for those training samples characterized by a prominent foreground and a simple background.

\vspace{5pt}
\noindent\textbf{Real-Mix versus Diff-Mix.} In Fig. \ref{fig:appendix_realmix_diffmix}, we compare the generated samples between Real-Mix and Diff-Mix. We observe that, by conditioning on the reference image, Real-Mix accurately captures the semantic meaning of the terminology name (see ``\texttt{chuck will widow}" in panel (a)). This feature of Real-Mix is consistent with its superior performance in few-shot classification, as depicted in Fig. \ref{fig: few-shot classification}. Additionally, we observe that Diff-Mix achieves more precise foreground editing (refer to Fig. \ref{fig:appendix_realmix_diffmix} (c) and (d)). This enhanced accuracy can be attributed to the class descriptor maintaining its focus on the semantic content without being diverted to other extraneous information.
\subsection{Experiments}
\label{sec:append_experiment}
\noindent\textbf{Few-shot classification in Pascal.}
\begin{figure}
    \begin{subfigure}{0.23\textwidth}
    \centering
    \includegraphics[width=1\linewidth]{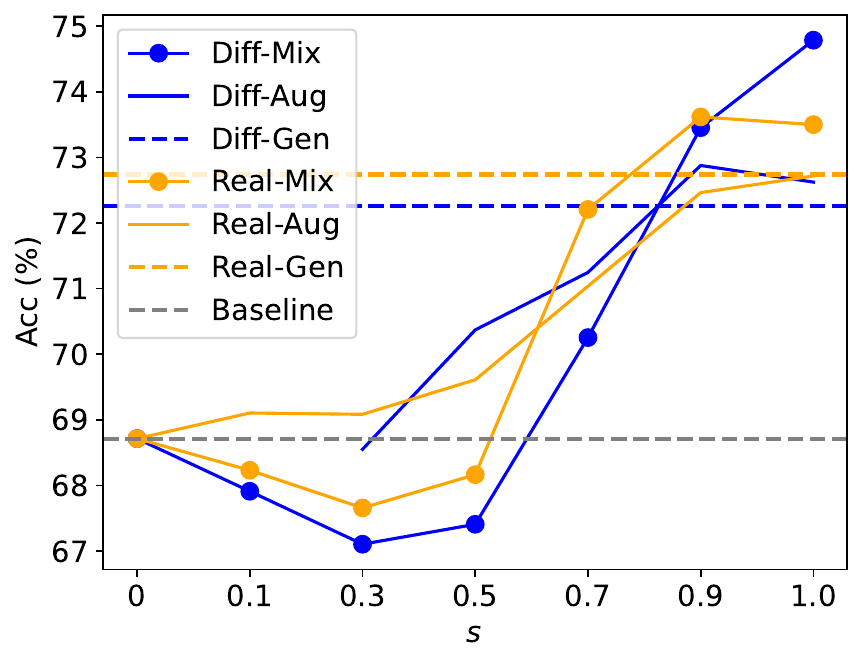}
    \caption{5-shot}
    \end{subfigure}
    \begin{subfigure}{0.23\textwidth}
    \centering
    \includegraphics[width=1\linewidth]{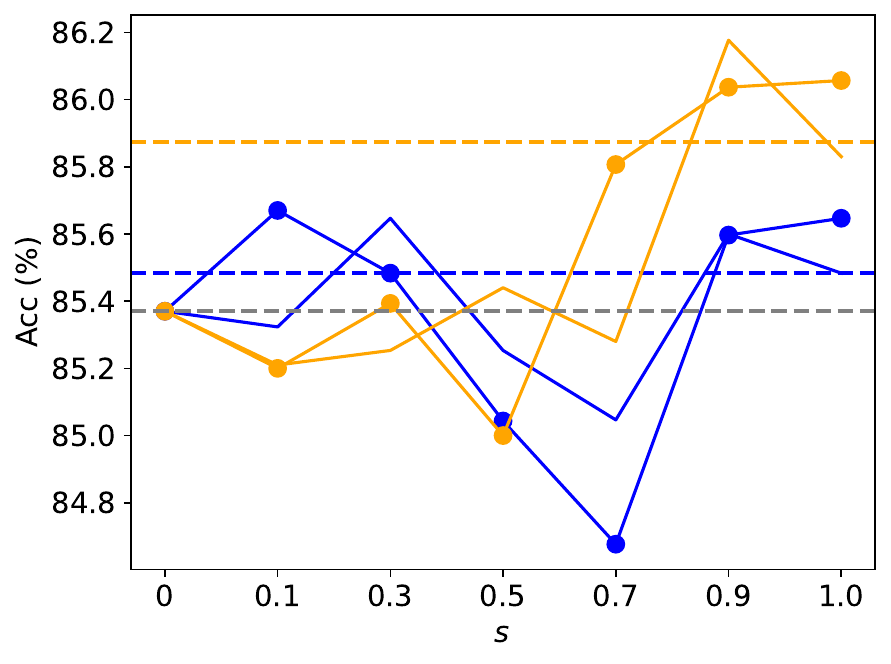}
    \caption{all-shot}
    \end{subfigure}
    \caption{5-shot and all-shot classification results in Pascal.}
    \label{fig:ex_pascal}
\end{figure}

\begin{table*}[th]
    \centering
    
    \footnotesize
    \begin{tabular}{l|ccc}\toprule
         hyperparameter            &  DB               & TI                & TI+DB                     \\\midrule
         Base Model                 &  Stable Diffusion-v1.5&Stable Diffusion-v1.5  & Stable Diffusion-v1.5    \\ 
         Optimized                  &  U-Net(LORA)            &      \texttt{[V$^i$]} & U-Net(LORA) + \texttt{[V$^i$]} \\
         Optimization Steps         &  35000                &     35000             & 35000                     \\
         Batchsize                  &  8                    &     8                 & 8                     \\
         Input Resolution           & 512$\times$ 512       & 512$\times$ 512       & 512$\times$ 512          \\
         Learning Rate              &  5e-5                 &    5e-5               & 5e-5                      \\
         Placeholder Token          &   -                   & \texttt{[V$^i$]}      & \texttt{[V$^i$]}          \\
         LORA Rank                  &        10             &      -                &  10                      \\\midrule
         \# if inference steps ($T$) &   25                  &       25              & 25                        \\
         Guidance Scale             &   7.5                 &       7.5             & 7.5                       \\
         Noise Scheduler            & DPMsolver++\cite{lu2022dpm} & DPMsolver++\cite{lu2022dpm}  & DPMsolver++\cite{lu2022dpm}  \\\bottomrule
         
    \end{tabular}
    \caption{Hyperparameters. This tables summarizes the hyperparameter settings of different fine-tuning strategies.}
    \label{tab:implementations_diffmix}
\end{table*}

In Sec. \ref{sec:appendix_visualization}, We have demonstrated that inter-class editing for general datasets tends to produce unrealistic images due to the visual gaps between two classes. Here, we present 5-shot and all-shot classification results in Figure \ref{fig:ex_pascal} for different expansion strategies on the general dataset Pascal \cite{pascal}. Originally, Pascal is an object class recognition dataset containing 11,530 images and 6,929 object segmentation masks. We construct it into a classification dataset, following the setting of Da-fusion \cite{dafusion}, resulting in a training split of 1,464 and a validation split of 1,449 for 20 general classes (\eg, cat and boat).
The main observation is that inter-class augmentation tends to be less effective for this general dataset, especially as the shot number increases (compare X-Aug and X-Mix in the figure). The effectiveness of fine-tuning also decreases, with Real-Gen consistently outperforming Diff-Gen. This suggests that the pre-trained SD is capable of generating sufficiently diverse and faithful samples for these coarse concepts.
Diff-Mix excels in handling domain-specific scenarios, where smaller differences in contour and layout between two classes are presented.

\subsection{Implementation Details.}
\label{sec:appendix_experiment_details}

\noindent\textbf{Diff-Mix.} Diff-Mix comprises two stages: the fine-tuning stage and the sampling stage. The implementation details of Diff-Mix for three different fine-tuning strategies are depicted in Table \ref{tab:implementations_diffmix}. Note that our fine-tuning strategy heavily relies on the diffuser \cite{von-platen-etal-2022-diffusers} repository. For DB and TI+DB, we only fine-tune the residual LORA matrices in attention modules in the U-Net. Please note that in the original Dreambooth \cite{ruiz2023dreambooth} paper, an unlearnable identifier was introduced to represent user-specific concepts in concept learning. However, in our implementation, we have opted not to use the identifier and have implemented it as a straightforward fine-tuning of text-to-image models. All fine-tuning and sampling processes are conduct on 4 RTX3090 GPUs. 

\vspace{5pt}
\noindent\textbf{Datasets.} We list the statistics of the datasets involved in our experiments in Table \ref{table:dataset}.

\begin{table}[]
    \centering
    \resizebox{0.98\linewidth}{!}{
    \begin{tabular}{l|cccl}\toprule
         Dataset          &  \# of classes & \# of training & \# of val. & Source  \\\midrule
         CUB              &   200          &  5994          & 5794       & \href{https://huggingface.co/datasets/Multimodal-Fatima/CUB_train}{Huggingface.co}         \\           
         FGVC Aircraft    &   100          &  3334          & 3333       & \href{https://huggingface.co/datasets/Multimodal-Fatima/FGVC_Aircraft_train}{Huggingface.co}          \\
         Oxford Flowers   &   102          &  4070          & 4119       & \href{https://huggingface.co/datasets/huggan/flowers-102-categories}{Huggingface.co}          \\
         Stanford Dogs    &   120          & 12000          & 8580       & \href{http://vision.stanford.edu/aditya86/ImageNetDogs/}{vision.stanford.edu}          \\
         Stanford Cars    &   196          & 8144           & 8041       & \href{https://huggingface.co/datasets/Multimodal-Fatima/StanfordCars_train}{Huggingface.co}          \\ \bottomrule
    \end{tabular}
    }
    \caption{Statistics of datasets.\label{table:dataset}}
    \label{tab:implementations_fewshot_model}
\end{table}


\begin{table}[]
    \footnotesize
    \centering
    \begin{tabular}{l|cc}\toprule
         hyperparameters & ResNet50 & ViT-B/16  \\\midrule
         Source          &   \href{https://github.com/pytorch/vision/blob/main/torchvision/models/resnet.py}{torchvision}        &   \href{https://github.com/pytorch/vision/blob/main/torchvision/models/vision_transformer.py}{torchvision}      \\               
         \# of parameters&  25.5M               &  86.6M            \\
         Pre-trained     &   ImageNet1K         &   ImageNet21K     \\                       
         Fine-tuned      &   -                  &   ImageNet1K      \\                       
         Input Resolution&   $448\times448$     &  $384\times384$   \\                           
         Batchsize       &   64                 &  32               \\
         Epochs          &   128                &  100              \\               
         Optimizer       &   SGD                &  SGD              \\                   
         Learning Rate   &   0.02               &  0.001            \\                       
         Momentum        &   0.9                &  0.9              \\                   
         Weight Decay    &   5e-5               &  5e-5             \\                       
         Label Smoothing &   0.9                &  0.9              \\\bottomrule
    \end{tabular}
    \caption{hyperparameters. This table summarizes the hyperparameter settings for CUB using two visual backbones in our conventional classification task.}
    \label{tab:implementations_model}
\end{table}

\vspace{5pt}
\noindent\textbf{Conventional classification.} The hyperparameter settings for the CUB dataset are presented in Table \ref{tab:implementations_model}. To reproduce Real-filtering (RF) \cite{SyntheticData}, a subset is derived from Real-Gen through data cleaning, as detailed in Section \ref{sec:appendix_dc}. Real-guidance (RG) \cite{SyntheticData} augments the dataset with low-strength intra-class editing, akin to Real-Aug with a strength parameter $s=0.1$. For the replication of Da-fusion \cite{dafusion}, we fine-tune the synthetic data (SD) using our TI strategy over 35,000 steps, with translation strengths randomly selected from the set ${0.25, 0.5, 0.75, 1.0}$. For CutMix and Mixup, the weight decay is $1 \times 10^{-5}$, and the mixup ratios are set to 0.1 and 0.3, respectively.

\vspace{5pt}
\noindent\textbf{Few-shot classification.} The few-shot classification is conducted on CUB with varying shot numbers: 1, 5, 10, and all. The comparison methods encompass: (1) inter-class augmentation strategies, namely Diff-Mix and Real-Mix, (2) intra-class augmentation strategies, namely Diff-Aug and Real-Aug, and (3) distillation-based methods, Diff-Gen and Real-Gen. The backbone model used is ResNet50 with an input resolution of $224^2$. We employ the same hyperparameters as in the conventional setting, as detailed in Table \ref{tab:implementations_model}, albeit with a larger batch size (256) and a higher learning rate (0.05). All experiments are conducted with three trials, and the average results are reported.

\begin{figure}
    \centering
\begin{minipage}{0.5\textwidth}
    \centering
    \resizebox{0.95\linewidth}{!}{
    \begin{tabular}{lccc}\hline
    Dataset           & \text{\#} of classes       &  \text{\#} of training     &  Imbalance Factor (IF) \\\hline 
    CUB-LT                & 200                 & \{1242, 1798, 2238\}           &    \{100,50,10\}           \\
    Flower-LT             & 102                 & \{847, 1238, 1532\}          &    \{100,50,10\}       \\\hline
    \end{tabular}
    }
    \captionof{table}{Statistics of long-tail datasets CUB-LT and Flower-LT.}
    \label{tab:longtail_statistics}
\end{minipage}
\end{figure}


\vspace{5pt}
\noindent\textbf{Long-tail classification.} Thanks to the authors of CMO \cite{cmo}, the reproduced long-tail results are built upon its open-source git repository \footnote{\href{https://github.com/naver-ai/cmo}{https://github.com/naver-ai/cmo}}. To construct the imbalanced dataset, an imbalance factor is introduced to control the imbalance level. The imbalance factor $\rho$ is defined as $\rho = \frac{\max_k \{n_k\}}{\min_k \{n_k\}}$, where $n_k$ is the number of samples in the $k$-th class. Specifically, given a normal dataset, we first sort the classes based on the number of images within classes in descending order, and use $k'$ to denote the sorted class index. A subset of images is randomly sampled from each class to achieve the desired imbalance, ensuring that the number of images for each class corresponds to the calculated target. The number of sampled images is determined by,
\begin{align}
    n_{k'} = \max\left(\bar{n} \times \left(\frac{1}{\rho}\right)^{\frac{k}{N-1}}, 1\right)
    \label{eq:longtail}
\end{align}
where $\bar{n}$ is the averaged number of images for each class, and $N$ is the total number of classes. 
The statistics of constructed CUB-LT and Flower-LT datasets can be found in Table \ref{tab:longtail_statistics}.
To uniformize the distribution of imbalanced real data, we first fix the number of iterative samples within each epoch and replace real samples with synthetic data with a 50\% probability. Note that the synthetic data conforms to the distribution specified by Eq. \ref{eq:longtail} with reversed class indices, thereby generating more synthetic images for tail classes. We maintain a constant number of training epochs and learning rate for both synthesis-free and synthesis-based approaches to ensure a fair comparison. We further present accuracy results for three distinct subsets when IF is 100: Many-shot classes (classes with over 20/30 training samples for CUB-LT/Flower-LT), medium-shot classes (classes with 5-20/10-30 samples for CUB-LT and Flower-LT), and few-shot classes (classes with fewer than 5/10 samples for CUB-LT/Flower-LT).



\subsection{Latency}
\label{sec:appendix_latency}
\begin{figure}
    \centering
    \begin{minipage}{0.5\textwidth}
    \includegraphics[width=1\linewidth]{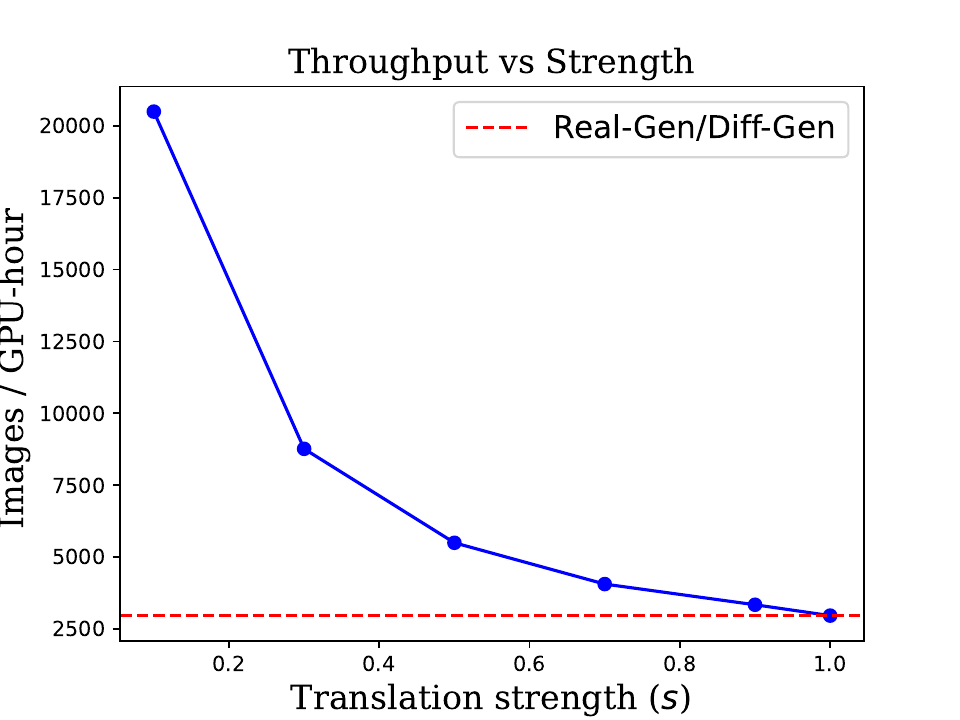}
    \caption{Sampling throughput across various translation strengths in a single RTX 3090 GPU. }
    \label{fig:throughput}
    \end{minipage}
    \begin{minipage}{0.5\textwidth}
    \vspace{1cm}
    \footnotesize
    
    \resizebox{0.95\linewidth}{!}{
    \centering
       \begin{tabular}{l|cc}\toprule
            Method&         Translation strength $s$ & Images per GPU-hour \\\midrule
            Real-filtering&     $\in$\{1.0\}                     & 2,957\\
            Real-guidance &     $\in$\{0.1\}                     & 20,502\\
            Da-fusion &         $\in$\{0.25,0.5,0.75,1.0\}       & 4,952     \\
            Diff-Mix &      $\in$\{0.5,0.7,0.9\}              & 4,179\\\bottomrule
            
       \end{tabular} }
       \captionof{table}{Comparison of sampling throughput of different expansion strategies.\label{table:throughput}}
    \end{minipage}
\end{figure}
Compared to non-generative augmentation methods, Diff-Mix's implementation incurs additional computational overhead during fine-tuning and data sampling. For instance, when working with the CUB dataset, which contains approximately 6,000 training samples, the fine-tuning process is completed in about 3 hours. This duration is achieved using an input resolution of $512 \times 512$ on 4 NVIDIA RTX 3090 GPUs with a total batch size of 8. During sampling, synthetic samples are generated at the same resolution with a total of $T=25$ reverse steps. The throughput across various translation strengths is evaluated in Fig. \ref{fig:throughput}, and a throughput comparison with other synthesis strategies is provided in Table \ref{table:throughput}. While Diff-Mix is more efficient than generating data from scratch, it is less so than low-strength editing (\eg, Real-guidance). For generating synthetic data for the CUB dataset with a 5x multiplier (resulting in approximately 30,000 images), the process requires roughly 2.5 hours using 4 NVIDIA RTX 3090 GPUs.

\section{Limitations}

Our inter-class augmentation method shows less effective when applied to general datasets that encompass a broad spectrum of concepts. We are optimistic, however, that integrating an image inpainting strategy or confining Diff-Mix to operate among adjacent classes could address this limitation. Moreover, the current annotation strategy is determined empirically and lacks a robust theoretical foundation, which may limit the generalizability of the strategy.


\end{document}